\documentclass[sigconf]{acmart}

\usepackage{booktabs} 
\usepackage{algorithm}
\usepackage[noend]{algpseudocode}

\usepackage{amsmath}
\usepackage{multirow}
\usepackage{amssymb}
\usepackage{mathtools}
\usepackage{afterpage}
\usepackage{pifont}
\newcommand{\cmark}{\ding{51}}%
\newcommand{\xmark}{\ding{55}}%
\usepackage{enumitem}

\usepackage{pgfplots}
\usepgfplotslibrary{dateplot, statistics}
\pgfplotsset{compat=1.12}

\algnewcommand\And{\textbf{and} }

\setcopyright{rightsretained}

\makeatletter
\newcommand*{\addFileDependency}[1]{
  \typeout{(#1)}
  \@addtofilelist{#1}
  \IfFileExists{#1}{}{\typeout{No file #1.}}
}
\makeatother

\acmDOI{10.1145/nnnnnnn.nnnnnnn}

\acmISBN{978-x-xxxx-xxxx-x/YY/MM}

\acmConference{KDD '20 (submitted)}{August 22--27}{San Diego, CA, USA}
\acmYear{2020}
\copyrightyear{2020}

\acmPrice{}

\begin{document}
	\title{ARMS: Automated rules management system for fraud detection}
	
	\author{David Apar\'icio}
    \email{david.aparicio@feedzai.com}
    \affiliation{%
      \institution{Feedzai}}
      
    	\author{Ricardo Barata}
    \email{ricardo.barata@feedzai.com}
    \affiliation{%
      \institution{Feedzai}}
      
      	\author{Jo\~ao Bravo}
    \email{joao.bravo@feedzai.com}
    \affiliation{%
      \institution{Feedzai}}
      
      	\author{Jo\~ao Tiago Ascens\~ao}
    \email{joao.ascensao@feedzai.com}
    \affiliation{%
      \institution{Feedzai}}
    
      	\author{Pedro Bizarro}
    \email{pedro.bizarro@feedzai.com}
    \affiliation{%
      \institution{Feedzai}}
      
      
    \renewcommand{\shortauthors}{D. Apar\'icio et al.}
      
	\begin{abstract}
		Fraud detection 
		is 
		essential 
		in financial services, with the potential of greatly reducing criminal activities and saving considerable resources for businesses and customers. We address online fraud detection, which consists of classifying incoming transactions as either legitimate or fraudulent in real-time. Modern fraud detection systems consist of a machine learning model 
		and rules defined by human experts. Often, the rules performance 
		degrades over time due to concept drift, especially of adversarial nature. Furthermore, they can be costly to maintain, either because they are computationally expensive or because they send transactions for manual review. We propose ARMS, an automated rules management system that evaluates the contribution of individual rules and optimizes the set of active rules using heuristic search and a user-defined loss-function. It complies with critical domain-specific requirements, such as handling different actions (e.g., accept, alert, and decline), priorities, blacklists, and large datasets (i.e., hundreds of rules and millions of transactions). We use ARMS to optimize the rule-based systems of two real-world clients. Results show that it can maintain the original systems' performance (e.g., recall, or false-positive rate) using only a fraction of the original rules ($\approx 50\%$ in one case, and $\approx 20\%$ in the other).
	\end{abstract}
	
	%
	%

	\ccsdesc[500]{Theory of computation~Optimization with randomized search heuristics}
	\ccsdesc[500]{Software and its engineering~Genetic programming}
	\ccsdesc[300]{Applied computing~Online banking}
	\ccsdesc[300]{Applied computing~Online shopping}
	\ccsdesc[300]{Applied computing~Secure online transactions}

	\keywords{fraud detection; genetic programming; evolutionary algorithms; greedy algorithms; randomized search}
	
	\settopmatter{printfolios=true}
	\maketitle
	
	\section{Introduction}
	
	\begin{figure*}[t]
		\centering
		\includegraphics[width=1\linewidth]{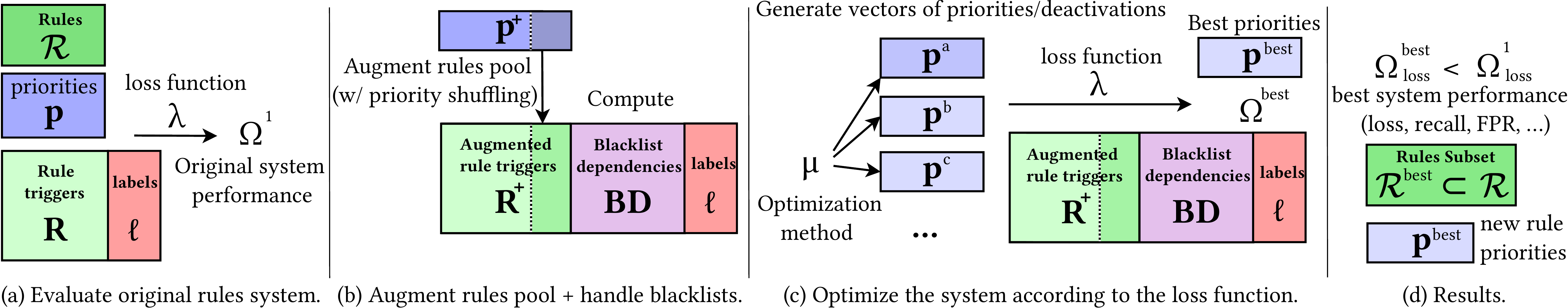}
		\vspace{-0.6cm}
		\caption{ARMS components: handling blacklists, priority shuffling, and optimizing a user-defined loss function.}
		\vspace{-0.3cm}
		\label{fig:arms_overview}
	\end{figure*}
	
	Financial institutions, merchants, customers, and government agencies alike suffer fraud-related losses daily, including credit card theft and other scams. Financial fraud consists of someone 
	inappropriately obtaining the details of a payment card (e.g., a credit/debit card) and using it to make unauthorized transactions. 
	Frequently, the cardholder detects such illicit usage and initiates a dispute with the bank to be reimbursed (a \emph{chargeback}), at the expense of the merchant or bank that accepted the transaction. 
	An over-conservative decision-maker might block all suspicious activity. However, this is far from optimal, as fraud patterns are not trivial, and it prevents legitimate economic activity. 
	Therefore, it is essential to adjust automated fraud detection systems to the risk profile of the client.
	
	
	Modern automated fraud detection systems consist of a machine learning (ML) model followed by a rule-based system. The model \emph{scores} the transaction. The rule-based system uses the score and triggers of manually defined rules to decide 
	an action 
	(i.e., accept, alert, or decline the transaction). Rule-based systems with many rules are complex, hard to maintain, and frequently computationally expensive. 
	An ideal system has only a minimum set of rules that ensure performance while preserving requirements and alerts low.
	
	Our main contributions are the following:
	\vspace{-0.05cm}
	\begin{enumerate}[leftmargin=2em]
	    \item Identifying a new problem: how to properly evaluate a complex rules system (taking into account overlapping rule triggers with different rule priorities and blacklists)? (Section~2).
	    \item Proposing ARMS (Figure~\ref{fig:arms_overview}), a framework which handles all bookkeeping necessary to correctly evaluate such rules systems (Sections~3.1--3.4).
	    \item Exploring optimization methods (namely random search, greedy expansion, and genetic programming) to improve the original system according to user-defined criteria (Sections 3.5-3.8).
	    \item Evaluating our proposed solutions on both synthetic and real data, demonstrating improvements to existing rules systems deployed at Feedzai (Section 4).
	\end{enumerate}
	\vspace{-0.05cm}

	Evaluating the performance of the whole fraud detection system is simple: given the fraud labels (i.e., the chargebacks) and the historical decisions, we compute performance metrics (e.g., recall at a given false positive rate or FPR). However, it is not enough to analyze the performance of each rule by itself. We need to consider how it contributes to the entire system as its triggers may overlap with other rules with different decisions and priorities. Blacklists 
	are another source of dependencies. Blacklisting rules, when triggered due to fraudulent behavior, blacklist the user (or email, or card) so that their future transactions are promptly declined. Deactivating blacklisting rules has side effects on the blacklists themselves and, therefore, in triggering or not rules that verify them. 
	
	In this work, we study the use case of a system with a pre-existing set of rules and priorities to optimize according to a user-defined objective function. As far as we know, we are the first to address the proper evaluation and optimization of such complex rules systems. A suitable goal is to minimize the number of rules and alerts while keeping the original system's performance (e.g., recall). 
	We explore three different methods (random, greedy, and genetic algorithms), using synthetic data and data sets from real-world online merchants.
	
	
	
	Our results show that ARMS can significantly reduce the number of rules while maintaining the system's performance. We stress that rules can depend on expensive aggregations (e.g., the average amount of the user's transactions in the last month). Thus, ARMS brings meaningful gains in practical fraud detection settings.
	
	We organize the remainder of the paper as follows. Section~\ref{sec:background} gives an overview of fraud detection systems 
	and discusses related work. Section~\ref{sec:arms} presents ARMS 
	main components: handling blacklists, rules system evaluation, priority shuffling, and rules system optimization. Section~\ref{sec:experiments} presents our results in synthetic data and real-world clients. Finally, we discuss our 
	conclusions in Section~\ref{sec:conclusions}.

	\section{Background}
	\label{sec:background}
	
	
	\subsection{Fraud detection}
	
	We focus on fraud detection in online payments, where a fraudster 
	makes unauthorized transactions 
	online. 
	Fraud detection can be formulated as a binary classification task: each transaction is represented as a feature vector, $\mathbf{z}$, and labeled as either \emph{fraudulent} (positive class, $y=1$) or \emph{legitimate} (negative class, $y=0$). Other approaches frame it as an outlier detection problem~\cite{kou2004survey} that treats fraudulent transactions as anomalies. Typical outlier detection is 
	unsupervised, 
	and often results in much lower performance.
	
	
	We consider fraud detection as a two-step process. First, when a transaction occurs, a feature engineering step, $g(\mathbf{z})$, is applied to the raw features $\mathbf{z}$, resulting in processed features, $\mathbf{x}$. An example of a processed feature (a \emph{profile}) is the number of transactions for a card in the last hour. Secondly, the automated fraud detection system evaluates the transactions and decides between three actions: to \emph{accept} the transaction, to \emph{decline} it, or to \emph{alert} it to be manually reviewed (so that specialized fraud analysts investigate it and produce a final decision). Reviews are complicated (i.e., subject to human error) and expensive, as they require specialized knowledge and introduce unnecessary friction for legitimate transactions.
	
	
	
	\subsection{Automated fraud detection system}
	
	We consider an automated fraud detection system consisting of a machine learning model followed by a rule-based system.
	
	\subsubsection{Machine learning model}
	
	The supervised machine learning model trains offline using historical data. 
	When evaluating a transaction, the model then produces a score, $\hat{y} \in \left[0,1\right]$, that is typically the probability of fraud given the features, $P \left(y=1 \mid \mathbf{x} \right)$.
	
	\subsubsection{Rule-based system}
	
	Rules consist of conditions and corresponding actions. Depending on the action, the rules can be \emph{accept}, \emph{alert}, or \emph{decline} rules. Rules may depend on the model score (e.g., if $\hat{y} < 0.5$ then accept the transaction), and the features (e.g., if the transaction is above a \emph{risky amount}, then alert/decline it). Since a transaction might trigger multiple rules with contradictory actions, priorities are necessary. Finally, rules can be switched on and off at any time. The rules system encapsulates all rules, their state (active or not), and priorities. Generally, the rule system is a function, $f(\mathbf{x}, \hat{y})$, that evaluates a list of rules and returns an action.
	
	\subsection{System evaluation}
	
	To assess system performance, we compare the system's decisions with the labels coming from chargebacks or the fraud analysts' decisions. Then, we compute the relevant performance metrics.
	
	\subsection{Rule evaluation}\label{sec:rule_evaluation}
	
	The rules system, $f(\mathbf{x}, \hat{y})$, receives the processed features and the model score and returns a decision to accept, alert, or decline. It comprises a set of rules, $\mathcal{R} = (R_1, R_2, \dots, R_k)$, applied \emph{individually} on incoming transactions. Hence, transactions may trigger none, one, some, or all of these rules.
	
\begin{figure}[t]
		\centering
		\includegraphics[width=1.0\linewidth]{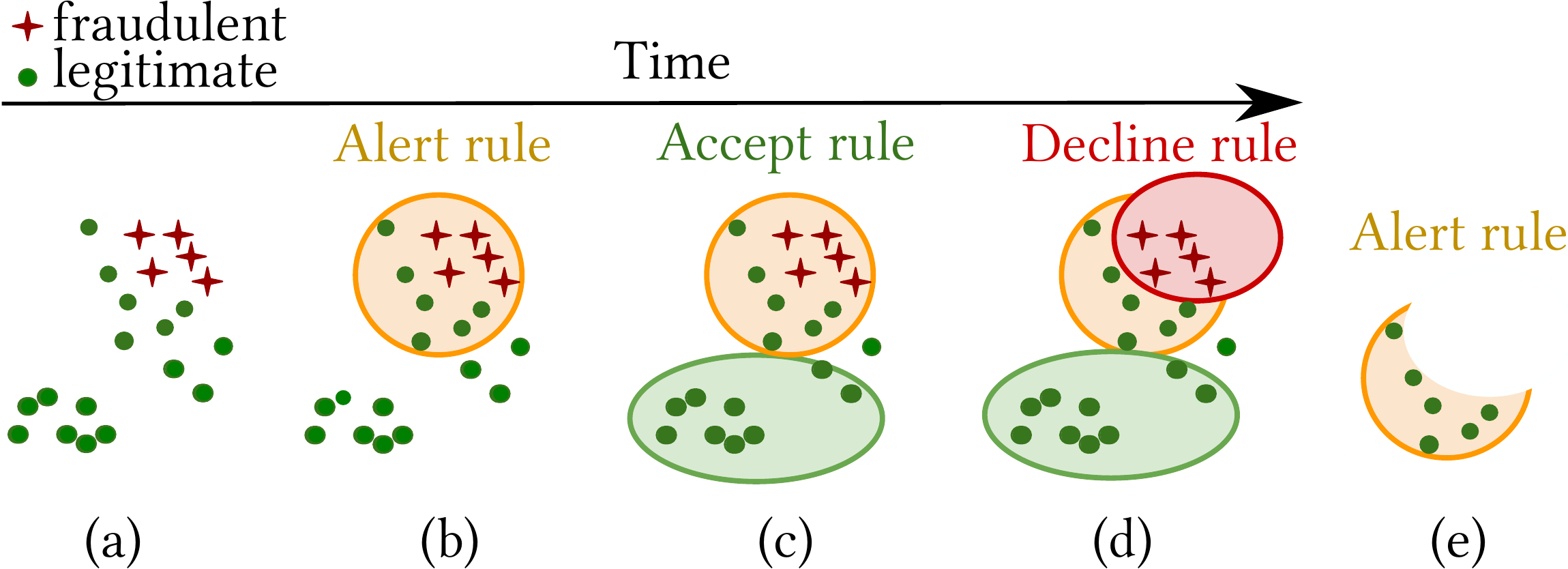}
		\vspace{-0.4cm}
		\caption{Rule degradation: (a) transactions in the feature space, (b-d) an alert, an accept, and a decline rule are progressively added, and trigger for some transactions. The alert rule has a good ratio of correct alerts when added, but by the end it is alerting only legitimate transactions (e).}
		\label{fig:rules_evolution}
		\vspace{-0.4cm}
	\end{figure}

	We aim to measure the contribution of individual rules to the system. Typically, at the time of the deployment of the system (i.e., after training with the latest data), rules and priorities perform well. However, as time goes by, fraud patterns change, and performance degrades. This degradation is 
	acute in fraud detection, given the 
	adversarial context (fraudsters often change their strategies).  Whereas some rules remain beneficial, others may become redundant or even degrade the performance of the system. Figure~\ref{fig:rules_evolution} illustrates how an initially good rule can degrade over time.
	As rule-based systems remain in production for a long time, it is essential to monitor how individual rules are impacting the system, namely their fraud detection and computational performances (rules can be heavy to compute, e.g., if they depend on profiles).
	
	One naive approach is to evaluate each rule independently by measuring how well its decisions match the labels (intuitively, accept rules should find legitimate transactions, while alert and decline rules should find fraudulent transactions). Then, if the rule's performance is inadequate, it is discarded. Notwithstanding, this approach is problematic and insufficient because it disregards interactions between rules. Consider the following examples:
	
	\begin{itemize}[leftmargin=2em]
	    \item Low-priority rules can perform outstandingly when no high-priority rules are triggered (e.g., in specific corner cases
	    ), but perform very poorly 
	    if used individually.
	    \item Turning off high-priority rules allows lower-priorities rules to act; this can lead to different decisions by the system. 
	\end{itemize}
	
	Instead of this naive approach, we build a rules management system that takes into account the interactions between rules with different actions and different priorities when evaluating them.
	
	\subsection{State-of-the-art}
	
	 \begin{table}[b]
\caption{Comparison of rules management systems.}
\vspace{-0.2cm}
\resizebox{0.48\textwidth}{!}{%
\scriptsize
\begin{tabular}{l|ccccccc}
             & \cite{ ishibuchi2004comparison} & \cite{allen1999using} & \cite{liu2015automated} & \cite{rosset1999discovery} & \cite{duman2011detecting} & \cite{gianini2020managing} & \textbf{ARMS} \\ \hline
             various rule actions   & \xmark & \xmark & \xmark & \xmark & \xmark & \xmark & \cmark\\
             rule priorities & \xmark & \xmark & \xmark & \xmark & * & \xmark & \textbf{\cmark}\\ 
             > million instances & \xmark & \xmark & \xmark & \xmark & \xmark & \xmark & \textbf{\cmark}\\ 
             user-defined loss & \xmark & \xmark & \xmark & \xmark & \xmark & \xmark & \textbf{\cmark} \\ 
             blacklists & \xmark & \xmark & \xmark & \xmark & \xmark & \xmark & \textbf{\cmark} \\ 
             rule learning & \xmark & \textbf{\cmark} & \xmark & \cmark & \xmark & \xmark & \xmark \\ \hline
             \multicolumn{4}{l}{* optimizes rule weights instead} & & & & \\
        \end{tabular}
        }
        \label{tab:system_survey}
        \vspace{-0.4cm}
\end{table}
	
	We review current work on the optimization of rule-based systems using search heuristics. Table~\ref{tab:system_survey} shows an overview of the methods.
	
	Ishibuchi et al. propose a method to maximize correctly classified instances, while reducing the number of rules, 
	using genetic programming 
	~\cite{ishibuchi1995selecting}. This approach is not sufficiently flexible for the fraud detection use-case, as the client (e.g., a merchant or bank) may want to optimize for other metrics (e.g., recall, or a combination of metrics)
	. Moreover, their method neglects priorities, and it is not clear if it scales up well for fraud detection data sets with millions of transactions (they used the Iris data set~\cite{fisher1936use}, which consists of 150 records).
	In a later study, they 
	compare multiple heuristics, namely greedy search and genetic programming, in four small data sets~\cite{ishibuchi2004comparison}.
	 
	Some approaches target specific use-cases, namely financial trading~\cite{allen1999using} or opinion mining~\cite{liu2015automated}.  Besides the domain, another crucial difference between our research and the work by Allen et al. is that, instead of learning new rules, we optimize existing rule systems.
	 
	
	Rosset et al. describe a method that learns and selects 
	rules for telecommunication fraud detection~\cite{rosset1999discovery}. Like us, the authors stress the importance of choosing a \emph{good} set of rules, instead of a set of \emph{good} rules. However, we target online transaction fraud and optimize a more complex system with priorities and blacklists. 
	
	Duman et al. propose a system combining genetic programming and scatter search to optimize rule \emph{weights} and other parameters~\cite{duman2011detecting}. Similar to our case, the rules are based on expert knowledge and suffer from concept drift. Each rule has a weight corresponding to its contribution to a fraud score, unlike our work, which considers priorities to activate a single rule. Furthermore, Duman et al. do not consider blacklists, different rule actions (e.g., accept, alert, and decline), and uses a predefined fitness function that minimizes the money loss. Additionally, the most substantial data set considered contains only $\approx 250$ thousand transactions and 43 rules and parameters. They report money savings of 212\% at the cost of a 67\% increase in false positives, and, after manual tuning, they settled for a system with savings of 189\% and a 35\% increase in false positives. 
	
	Gianini et al. optimize a system of 51 rules using a game theory approach~\cite{gianini2020managing}. They measure rule importance using Shapley values~\cite{winter2002shapley} as a measure of contribution to the system. They propose two strategies: (1) select the $n$ rules with highest Shapley values (and deactivate the others) and (2) greedy expansion of the set of rules using the Shapley values of the rules. Both strategies performed identically and were able to reduce the number of rules down to 30 while maintaining the original system's F-score. Like Duman et al. \cite{duman2011detecting}, this approach disregards essential constraints of the fraud detection system we are considering: rule priorities, rule actions, blacklists, and support for a user-defined loss function.
	
    

 \section{ARMS}
 \label{sec:arms}
	
	We start this section with an overview of ARMS. Then, we describe in detail each of its main components: handling blacklisting rules, the evaluation of the rule-based system, rule priority shuffle, and, finally, the optimization strategies to select rules. 
	
	
	\subsection{System overview}

	\begin{table}[t]
		\caption{Notation.}
		\vspace{-0.3cm}
		\label{tab:notation}
		\resizebox{\linewidth}{!}{%
			\begin{tabular}{l l}
				\toprule
				Features & $\mathcal{X} = (X_1, X_2, ..., X_m)$ \\
				Rules & $\mathcal{R} = (R_1, R_2, ..., R_k)$ \\
				Priority space & $\mathbb{P} = \{p \in \mathbb{Z} \mid p \geq -1\}$ \\
				Rule priority & $p_i \in \mathbb{P}$  \\ 
				Rule active condition & $p_i > -1$ \\
				Rules priority vector & $\mathbf{p} = (p_1, p_2, ..., p_k)$ \\
				Priority-action map & $a : p_i \rightarrow \{accept, alert, decline\}$ \\ \midrule
				Transaction feature vector & $\mathbf{x}$ = $(x_1, x_2, ..., x_m)$ \\
				Transactions & $\mathbf{X} = (\mathbf{x_1}, \mathbf{x_2}, ..., \mathbf{x_n})$ \\
				Transaction rules vector & $\mathbf{r} = (\{p_1, -1\}, \{p_2, -1\}, ..., \{p_k, -1\})$ \\ \midrule
				Rules triggers matrix & $\mathbf{R} = [\mathbf{r_{x_1}}, \mathbf{r_{x_2}}, ..., \mathbf{r_{x_n}}]^T$\\ 
				Labels vector & $\boldsymbol\ell = [\ell_{x_1}, \ell_{x_2}, ..., \ell_{x_n}]^T$ \\
				Blacklist updater rules & 
				$\mathcal{B}^u \subset \mathcal{R}$ 
				\\ 
				Blacklist checker rules & $\mathcal{B}^c \subset \mathcal{R}$  \\ 
				Loss function & $\lambda$ \\
				Performance & $\Omega$ (contains $\Omega_{loss}$, $\Omega_{recall}$, etc.)\\ \bottomrule
			\end{tabular}
		}
		\vspace{-0.5cm}
	\end{table}
	
	Algorithm~\ref{alg:arms} gives a general view of ARMS. We refer the reader to Table~\ref{tab:notation} for the notation used throughout this work.
	
	ARMS receives the following information as \textbf{inputs}:
	
	\begin{itemize}[leftmargin=2em]
		\item \textbf{Features}. A vector of features, $\mathcal{X}$ (e.g., username, email).
		\item \textbf{Transactions}. A matrix $\mathbf{X}^{n \times m}$ containing the values of the $m$ features for each of the $n$ transactions. It is needed to compute blacklists (i.e., to know blacklisted values for each feature, e.g., username = \emph{fraudster91}).
		\item \textbf{Triggers or activations}. A matrix $\mathbf{R}^{n \times k}$ containing the rule triggers of the $k$ rules for each of the $n$ transactions. Each cell $\mathbf{R}_{ij} = -1$ if rule $R_j$ did not trigger for transaction $\mathbf{x}_i$ or $\mathbf{R}_{ij} = p_j$ (i.e., the rule's priority) if it did.
		\item \textbf{Labels}. A vector with the label for each transaction, $\boldsymbol\ell$.
		\item \textbf{Priorities}. A vector with the priority of each rule, $\mathbf{p}$. 
		\item \textbf{Actions}. A map, $a$, mapping rule priorities to actions (i.e., accept, alert, or decline).
		\item \textbf{Blacklisting rules}. A set of blacklisting rules 
		containing rules that update the blacklist, $\mathcal{B}^u$, and rules that check it, $\mathcal{B}^c$.
		\item \textbf{Method}. Optimization strategy, $\mu$ (i.e., random search, greedy expansion, or genetic programming).
		\item \textbf{Loss function}. A loss function, $\lambda$, defined by the user.
		\item \textbf{Priority shuffle}. A boolean, $arp$, specifying whether to \underline{a}ugment the \underline{r}ules \underline{p}ool $\mathbf{R}$ by cloning rules with different priorities.  
		\item \textbf{Optimization parameters}. Set of parameters, $\theta$, which are specific to the optimization strategy (e.g., population size or mutation probability for the genetic algorithm or the number of evaluations for the random search).
	\end{itemize}

	ARMS starts by addressing the blacklist dependencies (line 1 of Algorithm~\ref{alg:arms}; details in Section~\ref{sec:blacklists}).
	Then, ARMS evaluates the original system's performance, $\Omega^{1}$ (line 2 of Algorithm~\ref{alg:arms}; Section~\ref{sec:evaluation}). This evaluation runs \emph{before} optimization because the loss function often depends on the original performance (e.g., optimize the FPR, while maintaining recall).
	Afterwards, ARMS augments the rules pool, if the user so desires (lines 3-4 of Algorithm~\ref{alg:arms};  Section~\ref{sec:priorityshuffle}). This adds new rules with the same triggers as existing rules, but with different priorities. The rationale is that changing priorities might improve the system.
	Finally, ARMS optimizes the rules system (line 5 of Algorithm~\ref{alg:arms};  Section~\ref{sec:optimization}). In essence, ARMS turns off rules and changes their priorities, obtaining a new priority vector, $\mathbf{p^{best}}$, to reduce the loss of the system, $\Omega^{best}$. 
	
	
	\begin{algorithm}
		\caption{ARMS: Automated Rules Management System.} 
		\label{alg:arms}
		\begin{flushleft}
			\textbf{Input:} Vector $\mathcal{X}$, Matrix $\mathbf{X}$, Matrix $\mathbf{R}$, vector $\boldsymbol\ell$, vector $\mathbf{p}$, map $a$, set $\mathcal{B}$, loss function $\lambda$, method $\mu$, parameter $arp \in \{0, 1\}$, parameters $\theta$\\
			\textbf{Output:} Vector $\mathbf{p^{best}}$, performance $\Omega^{best}$
		\end{flushleft}
		\begin{algorithmic}[1]
			\State $\mathbf{BD} \gets$ \Call{computeBlacklistDependencies}{$\mathbf{R}$, $\mathbf{X}$, $\mathcal{X}$, $\mathcal{B}$} 
			\State $\Omega^{1} \gets$ \Call{evaluate}{$\mathbf{X}$, $\mathbf{R}$, $\boldsymbol\ell$, $\mathbf{p}$, $a$, $\mathcal{B}$, $\mathbf{BD}$, $\lambda$} 
			\If{$arp$ = 1} 
			\State $\mathbf{R} \gets $ \Call{augmentRulesPool}{$\mathbf{R}$, $\mathbf{p}$, $a$}
			\EndIf
			%
			\State $(\mathbf{p^{best}}, \Omega^{best}) \gets$ \Call{$\mu$.optimize}{$\mathbf{X}$, $\mathbf{R}$, $\boldsymbol\ell$, $\mathbf{p}$, a, $\mathbf{BD}$, $\lambda$, $\Omega^{1}$, $\theta$}	
		\end{algorithmic}
	\end{algorithm}
	\vspace{-0.3cm}
	
	
	\subsection{Handling blacklists}\label{sec:blacklists}
	
	
	Both analysts and rules can blacklist entities. If an analyst finds transaction $\mathbf{x}$ to be fraudulent, they can blacklist some of its entities (e.g., in the future, always decline transactions from the email used in transaction $\mathbf{x}$). Similarly, \emph{blacklist updater rules} add entities to the blacklist when they trigger. Other rules, called \emph{blacklist checker rules}, trigger when a transaction contains a blacklisted entity.
	
	Therefore, blacklist rules have side effects. Deactivating blacklist updater rules can lead to blacklist checker rules not triggering, and affect the system's performance. Thus, we need to take this into account when evaluating the system. For this purpose, ARMS keeps a state of the blacklists and manages them according to the interaction between blacklist updater and blacklist checker rules (for a detailed description, we refer to Supplementary Algorithm~S\ref{alg:blacklists}).
	
	\subsection{Rules system evaluation}\label{sec:evaluation}
	
	ARMS evaluates (Algorithm~\ref{alg:evaluation}) the original system and the configurations  produced by the optimization strategies (Section~\ref{sec:optimization}). It creates an empty confusion matrix, $\mathbf{V}$ (line 2), to be updated by traversing each transaction, $\mathbf{x} \in \mathbf{X}$, alongside its rule triggers, $\mathbf{r_i} \in \mathbf{R}$, and its label, $\ell_{i} \in \boldsymbol\ell$ (lines 3-9). For each transaction:
	
	\begin{enumerate}[leftmargin=2em]
		\item ARMS computes the activations $\mathbf{r'_i}$ (i.e., what rules are active and with what priority), using priority vector $\mathbf{p}$ (line 4). When ARMS is evaluating the original system, $\mathbf{p}$ contains the original rules' priorities; however, rule priority shuffling and optimization strategies generate variations of $\mathbf{p}$. 
		\item ARMS checks whether to turn off any blacklist checker rules as a side-effect and stores that in $\mathbf{r''_i}$ (line 5). 
		\item ARMS obtains the final decision, $o_i$, from $\mathbf{r''_i}$, i.e., to accept, alert, or decline (line 6). It is the action of the highest priority rule triggered for the transaction that is active.
		\item ARMS evaluates the decision, $o_i$, against the label $\ell_i$, storing it in $v_i$ (line 7). Accepting a legitimate transaction is a true negative. Declining/alerting a legitimate transaction is a false positive. Declining/alerting a fraudulent transaction is a true positive. Accepting a fraudulent transaction is a false negative. The confusion matrix, $\mathbf{V}$, is updated with $v_i$ (line 8).
	\end{enumerate}

	Finally, ARMS uses the confusion matrix $\mathbf{V}$ to compute the rule configuration's performance, $\Omega$, based on a user-defined loss function, $\lambda$ (line 9). The loss function allows optimizing (e.g., minimize the number of active rules, maximize recall) and satisfying metrics or constraints (e.g., keep the original system's FPR). We discuss loss functions used in synthetic data and real-world clients in Section~\ref{sec:experiments}.
	
	\begin{algorithm}
		\caption{Rules system evaluation.} 
		\label{alg:evaluation}
		\begin{algorithmic}[1]
			\Function{evaluate}{$\mathbf{X}$, $\mathbf{R}$, $\boldsymbol\ell$, $\mathbf{p}$, $a$, $\mathcal{B}$, $\mathbf{BD}$, $\lambda$}
			\State $\mathbf{V} \gets $ \Call{initConfusionMatrix}{{}}
			\ForAll{$\mathbf{x} \in \mathbf{X}, \mathbf{r_i} \in \mathbf{R}, \ell_i \in \boldsymbol\ell$}
			\State $\mathbf{r'_i} \gets$ \Call{mask}{$\mathbf{r_i}$, $\mathbf{p}$}
			\State $\mathbf{r''_i} \gets$ \Call{handleBD}{$\mathbf{r'_i}$, $\mathcal{B}$, $\mathbf{BD}[\mathbf{x}]$}
			\State $o_i \gets$ $a$(\Call{max}{$\mathbf{r''_i}$})
			\State $v_i \gets$ \Call{getTruthValue}{$o_i, \ell_i$}
			\State $\mathbf{V} \gets$ \Call{updateConfusionMatrix}{$\mathbf{V}$, $v_i$}
			\EndFor
			\State $\Omega \gets$ \Call{$\lambda$}{$\mathbf{V}$} 
			\State \Return $\Omega$
			\EndFunction
		\end{algorithmic}
	\end{algorithm}
	
	\vspace{-0.3cm}
	
	
	\subsection{Priority shuffle}~\label{sec:priorityshuffle}
	
	Initial rule priorities require expert knowledge and are defined by clients or fraud analysts
	. Over time, however, the system requires adjusted priorities to deal with concept drift and incorporate emerging knowledge (e.g., new rules). In this section we discuss how ARMS addresses priority shuffling for optimization (Section~\ref{sec:optimization}). First, we discuss how ARMS changes the priority of individual rules. Then, we discuss how ARMS can augment the initial rules pool by cloning existing rules and assigning them alternative priorities.
	
	\subsubsection{Random priority shuffle} 
	
	%
	%

	Since the system might have many rules and many possible priorities, the search space of all possible rule priorities can be gigantic. A more efficient alternative for such cases is to use random priority shuffle.	For a given rule $r_i$ with priority $p_i$, ARMS changes its priority to $p_j \not\eq p_i$ with the same action, i.e., $a_i = a_j$. The new rule priority is sampled considering uniform probabilities.
	Consider the illustrative example with three types of accept rules: \emph{weak accept} with priority 1, \emph{strong accept} with priority 3, and \emph{whitelist accept} with priority 5. Random priority shuffle can, for example, change the priority of a strong accept to either 1 (weak accept) or 5 (whitelist accept).

	\subsubsection{Augment rules pool} 
	
	%

	Another option is to augment the initial pool by cloning existing rules (i.e., same triggers), but assigning them different priorities. Starting from the existing priorities, $\mathbf{p}$, we create variants for each $p_i \in \mathbf{p}$, 
	with all possible alternative priorities with the same action, $\mathbf{E}$.
	Then, for each $p_j \in \mathbf{E}$, ARMS adds a new vector (a new "rule")
	with the same triggers as the original rule and the new priority, $p_j$, to the rules triggers matrix, $\mathbf{R}$.
	
	\subsection{Optimization strategies}\label{sec:optimization}
	
	ARMS uses two fundamental mechanisms to optimize a rule-based system: deactivate 
	underperforming rules and change priorities. It is unfeasible to test all possible combinations. Instead, we employ three heuristics (methods): random search (Section \ref{sec:randomsearch}), greedy expansion (Section \ref{sec:greedyexpansion}), and genetic programming (Section \ref{sec:geneticprogramming}).

	First, we give an overview of ARMS optimization (Algorithm~\ref{alg:arms_optimization}), as methods share a similar structure. The original system (i.e., with rule priorities $\mathbf{p}$ and performance $\Omega^1$) is the one to beat (line 1). Until meeting a predefined stopping criteria (line 3), ARMS generates new priority vectors, $\mathbf{p'}$, which are variations of the original $\mathbf{p}$ (line 4). The criteria can be to stop after $k$ hours, after computing $n$ variations, or when the loss between consecutive iterations does not improve above a threshold $\epsilon$.  ARMS saves the variation with lowest loss it finds, $\mathbf{p^{best}}$,  alongside its performance $\Omega^{best}$, and returns them to the user (lines 5-8). The fundamental difference between methods is how they generate the variations, $\mathbf{p'}$.
	
		\begin{algorithm}
		\caption{ARMS optimization.} 
		\label{alg:arms_optimization}\begin{flushleft}
			$\theta$: parameters of the method
		\end{flushleft}
		
		\begin{algorithmic}[1]
			\Function{$\mu$.optimize}{$\mathbf{X}$, $\mathbf{R}$, $\boldsymbol\ell$, $\mathbf{p}$, $a$, $\mathbf{BD}$, $\lambda$, $\Omega^{1}$, $\theta$}
			\State ($\mathbf{p^{best}}, \Omega^{best}) \gets  (\mathbf{p}, \Omega^{1})$
			\While{\Call{stoppingCriteriaNotMet}{{}}}
				\State generate a new $\mathbf{p'}$ from $\mathbf{p}$
				\If{$\mathbf{p'}$ is the best so far}
					\State save it as $\mathbf{p^{best}}$
					\State save its performance as $\Omega^{best}$
				\EndIf
			\EndWhile
			\State \Return ($\mathbf{p^{best}}, \Omega^{best}$)
			\EndFunction
			
		\end{algorithmic}
	\end{algorithm}	
	
	\subsection{Random search}\label{sec:randomsearch}
	
	A straightforward approach is to generate random rules priority vectors, $\mathbf{p}'$, and evaluate them against the original $\mathbf{p}$, saving the best rule configuration $\mathbf{p}'$ that it found. While this approach seems naive, it is a natural baseline that can be better and less expensive than grid or manual searches~\cite{bergstra2012random}. Random search has two parameters: 
	
	\begin{itemize}[leftmargin=2em]
		\item \emph{Rule shutoff probability}, $\rho$. Percentage of rules to deactivate, e.g., if $\rho = 50\%$, then ARMS turns off $\approx 50\%$ of the rules.
		\item \emph{Rule priority shuffle probability}, $\gamma$. Percentage of rules with priorities changed, e.g., if $\gamma = 50\%$, then ARMS generates new priority vectors for $\approx 50\%$ of the rules.
	\end{itemize}
	
	For more detail, we refer to Supplementary Algorithm~S\ref{alg:random}.
	
	\subsection{Greedy expansion}\label{sec:greedyexpansion}
	
	 ARMS contains a greedy expansion module, that starts from a set of inactive rules and greedily turns on rules, one at the time.
	 Greedy solutions are not guaranteed to find the global optimum. 
	Consider the following example, where we want to optimize recall and rules $R_1$, $R_2$, and $R_3$ have recall 70\%, 69\%, and 20\%, respectively. A greedy solution would pick $R_1$ first. Now, imagine that rules $R_2$ and $R_3$ are detrimental to $R_1$, i.e., the system becomes worse if we combine $R_1$ with either $R_2$ or $R_3$. Hence, the final solution is a system with only $R_1$. Imagine, however, that $R_2$ and $R_3$ are somewhat complementary, and that, when combined, the system's recall is $>70\%$. Then, the global optimum is $>70\%$, and the greedy solution is not optimal. Nevertheless, greedy heuristics can find useful solutions in a reasonable time. 
	
	
	
	For more detail, we refer to Supplementary Algorithm~S\ref{alg:greedy}.
	
	\subsection{Genetic programming}\label{sec:geneticprogramming}
	
	Genetic programming is standard in classification tasks~\cite{espejo2009survey}, such as fraud detection. It continuously improves a population of solutions by combining them using crossovers and random mutations, while keeping a fraction of the best solutions for the next iteration. 
	
	In our case, we build a population of random rule configurations and improve them with genetic programming. The algorithm has three parameters:
	
	\begin{itemize}[leftmargin=2em]
		\item \emph{Population size}, $\psi$. Number of configurations per iteration, e.g., if $\psi = 100$, ARMS evaluates 100 different rule configurations per iteration.
		\item \emph{Survivors fraction}, $\alpha$. Fraction of the top configurations that survive for the next iteration, e.g., if $\psi = 100$ and $\alpha = 20\%$, only the 20 best solutions \emph{survive} for the next iteration. If $\alpha$ is high, then we might achieve higher variability but get stuck trying to improve bad solutions. If $\alpha$ is low, then the lack of variability might prevent the system from reaching a good solution. 
		\item \emph{Mutation probability}, $\rho$. The percentage of rules subject to random mutation, e.g., if $\rho = 20\%$, then 20\% of the rules are randomly mutated (i.e., the \emph{child} rule configuration mutates the \emph{parents} rules configuration). If $\rho$ is high, we leave little room for genetic optimization and are essentially doing a random search. If $\rho$ is low, we are more dependent on finding good parent configurations. 
	\end{itemize}
	
	For more detail, we refer to Supplementary Algorithm~S\ref{alg:genetic}.

	\section{Experiments and results}
	\label{sec:experiments}
	
	We test the following hypotheses: \textbf{(h1)} ARMS turns off rules and, at least, maintains system performance, \textbf{(h2)} ARMS changes the priority of rules and improves system performance, \textbf{(h3)} results are stable (i.e., similar across folds).
	
	 \subsection{Synthetic data}
	
	Since we can not find public data sets similar to our own, we use synthetic data to test hypotheses \textbf{(h1--h2)}. Later, we also test \textbf{(h1--h3)} in real datasets. 
	We generate 225k labels with a fraud rate of $5\%$ (i.e., 11250 positive labels) and simulate accept, alert, and decline rules from the labels. The support of a rule corresponds to how many times it triggers. An accept rule with negative predictive value (NPV) of $k\%$ is correct $k$\% of the times that it triggers (i.e., out of all triggers, $k$\% will be true negatives). The same goes for the precision (PPV) of an alert or decline rule (i.e., out of all triggers, $k$\% will be true positives). We sample the support, NPV and precision from Gaussian distributions and use 10 different priority levels (for details see Supplementary Section~\ref{sup_sec:synthetic}) and divide the data set into three splits: train, validation, and test, with 75k "transactions" each.
	
	
	\subsection{Methodology}
	
	We run ARMS 
	on the train set and do parameter tuning 
	in the validation set. We detail the parameter space in Supplementary Section~\ref{sec:hyperparams}. We ensure that results are comparable between random search and genetic programming by keeping the number of rule configuration evaluations fixed (i.e., $n = 300$k). 
	
	We optimize the loss function from 
	Equation~\ref{eq:loss_synthetic} with $\alpha = 0.1$, $\beta = 0.5$, and $\gamma = 0.4$. Note that $\Omega^1$ and $\Omega'$ are the performance of the original system and of a configuration found by ARMS, respectively; $\Omega_{rules\%}$ is the percentage of active rules, $\Omega_{recall\%}$ is the recall, and $\Omega_{alert\%}$ is the alert rate.
	
	\vspace{-0.15cm}
    \begin{equation}
    	\label{eq:loss_synthetic}
    	    \lambda(\mathbf{R'}) = 
        	    \alpha * \Omega^{'}_{rules\%} - \beta * \Omega^{'}_{recall} + \gamma * \Omega^{'}_{alerts\%}
	\end{equation}
	\vspace{-0.2cm}
	
	Finally, we evaluate the four final methods in the test set: the original system, and the best rule system configuration found by random search, greedy expansion, and genetic programming. 
	
	\subsection{Results on synthetic data}
	
	After running parameter tuning (note that the greedy expansion method does not have any parameter), 
	we find that the following parameters were the best:
	
	\begin{itemize}[leftmargin=2em]
	    \item \textbf{Random search: } $\rho = 40\%$.
	    \item \textbf{Genetic programming: } $\rho = 10\%$, $\psi = 30$, $\alpha = 5\%$.
	\end{itemize}
	
For brevity, we omit results for other parameters; we do a more thorough analysis of the parameters in real data sets (Section~\ref{sec:results_real_data}).  We observe that all methods improved upon the original system, and that genetic programming was the one with highest performance (Table~\ref{tab:synthetic_dataset_results}). When we ran augmented rules pool (ARP) before optimization, results consistently improved. Thus, 
 we verify hypothesis \textbf{(h1)} and \textbf{(h2)}.

		\begin{table}[b]
	    \centering
	    \caption{Performance of ARMS in synthetic data.}
	   \vspace{-0.25cm}
	    \label{tab:synthetic_dataset_results}
	    \begin{tabular}{l|cccc}
	         & \textbf{recall} & \textbf{alerts \%} & \textbf{rules off} & \textbf{loss} \\ \hline
	         \textbf{original} & 13.11\% & 0.779\% & none & 0.0376 \\
	         \textbf{random} & 79.53\% & 1.013\% & 38 (38.8\%) &  -0.1837 \\
	         \textbf{greedy} & 54.42\% & 1.746\% & 34 (34.7\%) & -0.1998 \\
	         \textbf{genetic} & 52.82\% & 1.067\% & 45 (45.9\%) & -0.2058 \\
	         \textbf{greedy w/ arp} &  53.30\% & 1.107\% & 43 (43.9\%) & -0.2060 \\
	         \textbf{genetic w/ arp} &  53.09\% & 0.97\% & 45 (45.9\%) & \textbf{-0.2075} \\
	    \end{tabular}
	\end{table}

	\subsection{Real-world data sets}
	\label{sec:datasets}
	
	We evaluate ARMS on representative samples of real-world data sets of two online merchants. In both cases, an automated fraud detection system actively scores transactions in production. We collected the rule triggers, model decisions, and blacklists. The data sets comprise dozens of rules, with different actions (i.e., accept, alert, and decline) and multiple priorities (
	more details in Supplementary Section~\ref{sup_section:realworlddatasets}). For privacy compliance, we refer to the data sets simply as \textbf{D1} and \textbf{D2}.
    
    The data covers six months of transactions. We divide each data set in four sequential and overlapping folds of three months each (for temporal cross-validation, detailed in Section~\ref{sec:methodology_tcv}) and split each fold into three sequential sets (train, validation, and test) of one month each.
	
\begin{figure}
		\centering
		\includegraphics[width=0.9\linewidth]{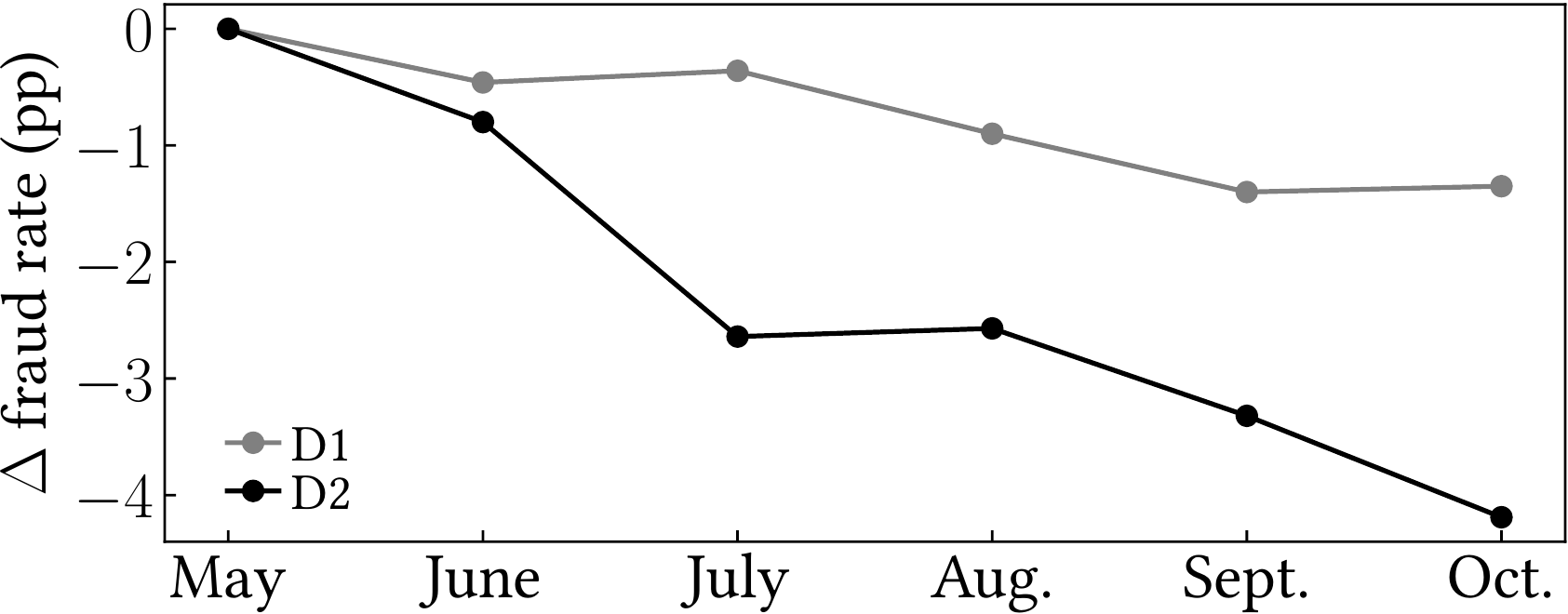}
		\vspace{-0.2cm}
		\caption{Datasets fraud rate evolution (concept drift).}
		\label{fig:datasetb_fraudrate}
	\end{figure}

	Unless explicitly stated, when we mention \emph{fraud}, we are referring to validated fraud (i.e., chargebacks or fraud confirmed by analysts, not transactions declined by the automated fraud detection system). Due to the adversarial setting and other factors, we observe concept drift in both data sets. Figure~\ref{fig:datasetb_fraudrate} shows the evolution of the fraud rate in \textbf{D1} and \textbf{D2} (with May 2018 as reference), highlighting the system's ability to reduce fraud over time. 
	
	While both clients are online merchants, they have three important differences:
	
	\begin{enumerate}
	    \item \textbf{D1} has more non-verified declined transactions. It has $\approx$ 14x more auto-declined transactions than confirmed frauds, due to the specific requirements of the client. Using automatically declined transactions for training is dangerous as it creates a feedback loop. Thus, we disregard them in training and validation but use them in testing so that results are comparable to a production setting. Moreover, for this dataset, ARMS does not optimize decline rules. 
	    \item Only \textbf{D2} uses blacklists.
	    \item The active rules in \textbf{D2} changed multiple times during the period under study, while the rules in \textbf{D1} never changed.
	\end{enumerate}
%
	
	\subsection{Methodology}
	\label{sec:methodology}
	
	\subsubsection{Optimization metrics (loss functions)}
	
	Online merchants are required to keep the fraud-to-gross rate (FTG) under a certain threshold, or else they face fines.
	Thus, a sensible approach is to minimize the FPR and ensure that recall is within the legal requirements. The system should be able to pick up all the necessary fraud (ideally, all of it) without declining legitimate transactions. Additionally, reducing the number of rules and alerts decreases the overall cost of the system. We use different loss functions for each data set, showing ARMS' ability to fit diverse use-cases:
	
	\begin{itemize}
    	\item In \textbf{D1}, the FPR is artificially high due to the many transactions declined by the automated fraud detection system. Therefore, our focus is to remove rules, $\Omega'_{rules\%}$, and reduce alerts, $\Omega'_{alert\%}$, while maintaining approximately the same recall, $\Omega'_{recall}$, as the original rule-based system, $\Omega^1_{recall}$, (Equation~\ref{eq:loss_dataseta}). We use $\alpha = \beta = \frac{1}{2}$, thus giving equal importance to both objectives. 
    	
    	
    	\item In \textbf{D2}, the objective is to remove rules, $\Omega'_{rules\%}$, but also to improve recall, $\Omega'_{recall}$, while maintaining approximately the same FPR, $\Omega'_{fpr}$, as the original system, $\Omega^1_{fpr}$, (Equation~\ref{eq:loss_datasetb}). We use $\alpha = 0.05$ and $\beta = 0.95$, thus attributing more importance to improving recall than to reducing the number of rules. 
	\end{itemize}
	
	\begin{equation}
    	\label{eq:loss_dataseta}
    	    \lambda(\mathbf{R'}) = 
        	    \begin{cases} \alpha * \Omega^{'}_{rules\%} + \beta * \Omega^{'}_{alert\%} & \text{if } \Omega^{'}_{recall} \geq 0.95* \Omega^{1}_{recall} \\
        	    \alpha + \beta + (\Omega^{1}_{recall} - \Omega^{'}_{recall}) & \text{otherwise}
        	    \end{cases}
	\end{equation}
	
	\begin{equation}
    	\label{eq:loss_datasetb}
    	    \lambda(\mathbf{R'}) = 
        	    \begin{cases} \alpha * \Omega^{'}_{rules\%} - \beta * \Omega^{'}_{recall} & \text{if } \Omega^{'}_{fpr} \leq \Omega^{1}_{fpr} \\
        	    \alpha + (\Omega^{1}_{fpr} - \Omega^{'}_{fpr}) & \text{otherwise}
        	    \end{cases}
	\end{equation}
	
	
	\bigbreak
	
	\subsubsection{Baselines}
	
	We compare ARMS optimized rule systems against three baselines:
	
	\begin{enumerate}
	    \item \textbf{Original system (\emph{All on}): } system with all rules and original priorities.
	    \item \textbf{Mandatory system (\emph{All off}): } system with no rules except for the ones that cannot be deactivated due to business reasons, with the original priorities.
	    \item \textbf{Random search: } generate $r$ independent rule configurations, using different values of $\rho$ (Section~\ref{sec:randomsearch}).
	\end{enumerate}
	
	If ARMS finds rule systems better than baselines 1 and 2 by turning off rules, we successfully address \textbf{(h1)}. If it further improves its performance by also tuning rule priorities, we  address \textbf{(h2)}.
	
		\subsubsection{Temporal cross validation (TCV)}
	\label{sec:methodology_tcv}
	
	We use TCV to verify \textbf{(h3)}. For each data set, we create four folds composed of three sets (i.e., train, validation, and test) of one month each. We train ARMS with different search heuristics and parameters on each train data set, evaluate the resulting configurations on the validation data set, and identify the best one for each heuristic. Finally, we evaluate the winners and the baselines on the test set.
	
	\subsubsection{Optimization strategies}
	
	We run ARMS with two different optimization strategies: greedy expansion (Section~\ref{sec:greedyexpansion}) and genetic programming (Section~\ref{sec:geneticprogramming}). Results for both are shown in Sections~\ref{sec:greedy_real} and \ref{sec:genetic_real}, respectively.
	     
	 \subsection{Results on real data}
	 \label{sec:results_real_data}
	 
	    Unless stated otherwise, the results refer to rule configurations obtained in the train data of each fold and evaluated in the respective validation set. 
	    Results shown are always relative to the original system baseline and show the gains relative to the current system in production, i.e., $\Delta loss$ is the difference between the loss of the system being evaluated and the original one.
	    
	    \subsubsection{Baselines comparison}
	    
\begin{figure}
    		\centering
    		\includegraphics[width=1.0\linewidth]{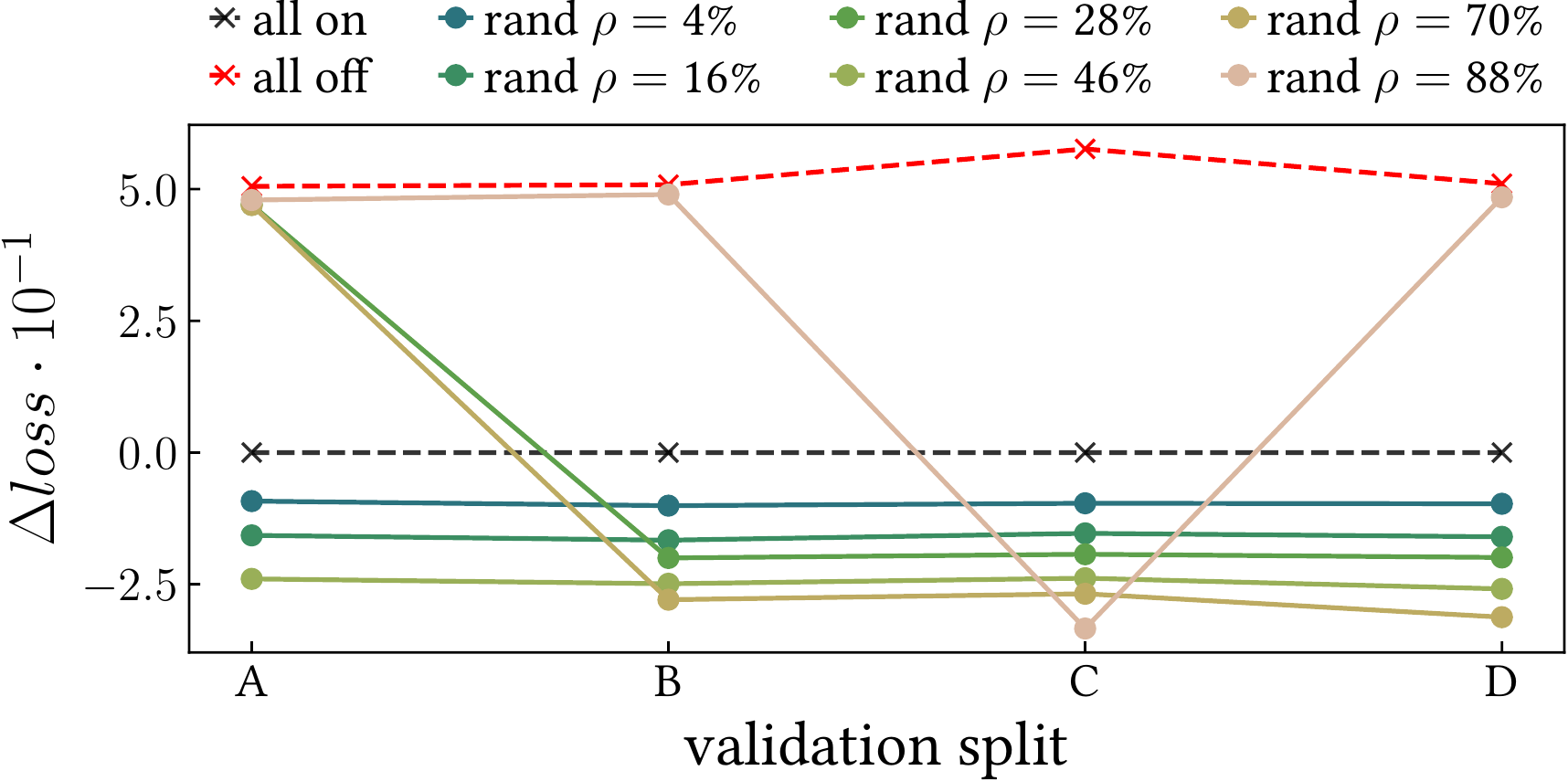}
    		\vspace{-0.6cm}
    		\caption{Baselines comparison in \textbf{D1}.}
    		\label{fig:dataseta_baselines}
	    \end{figure}

	    We compare the original system (\emph{all on}) against the mandatory system (\emph{all off}) and against random search, with $n = 10000$ and $\rho$ as a tunable parameter with values spaced out in 4\% intervals (Figure~\ref{fig:dataseta_baselines} for \textbf{D1}
	    ). We observe that the mandatory system has a higher loss than the other systems, as it fails to meet the recall constraint from Equation~\ref{eq:loss_dataseta}. We also observe that random search is almost always superior to the original system, regardless of $\rho$. In a few cases, the random search is worse than the original system because it does not meet the recall constraint, namely with aggressive configurations (e.g., $\rho = 88\%$). On the other hand, aggressive random search (higher $\rho$) can decrease the loss significantly, so there is a trade-off between being able to meet the recall constraints and lowering the loss. We observe similar behavior for \textbf{D2}, and thus omit results for brevity. 
	    Nevertheless, we show metrics besides the loss for \textbf{D2} in Supplementary Figure~S\ref{fig:datasetb_baselines}. From these results, we decide to use random search with $\rho = 46\%$ 
	    for \textbf{D1}, and $\rho = 58\%$ 
	    for \textbf{D2}, for the baselines, alongside the original system and the mandatory system for both data sets.
	    
	    \subsubsection{Greedy expansion results}
	    \label{sec:greedy_real}
	    
	    We test greedy expansion with and without ARP. We find that ARP did not improve the system in \textbf{D1} or \textbf{D2}. 
	    One possible explanation is that greedy expansion yields simple systems with few rules, so it did not benefit from ARP. 
	    Another possibility is that the original priorities are already well-tuned for both data sets as they correspond to mature systems. 

	    \begin{figure}[t]
    		\centering
    		\includegraphics[width=1.0\linewidth]{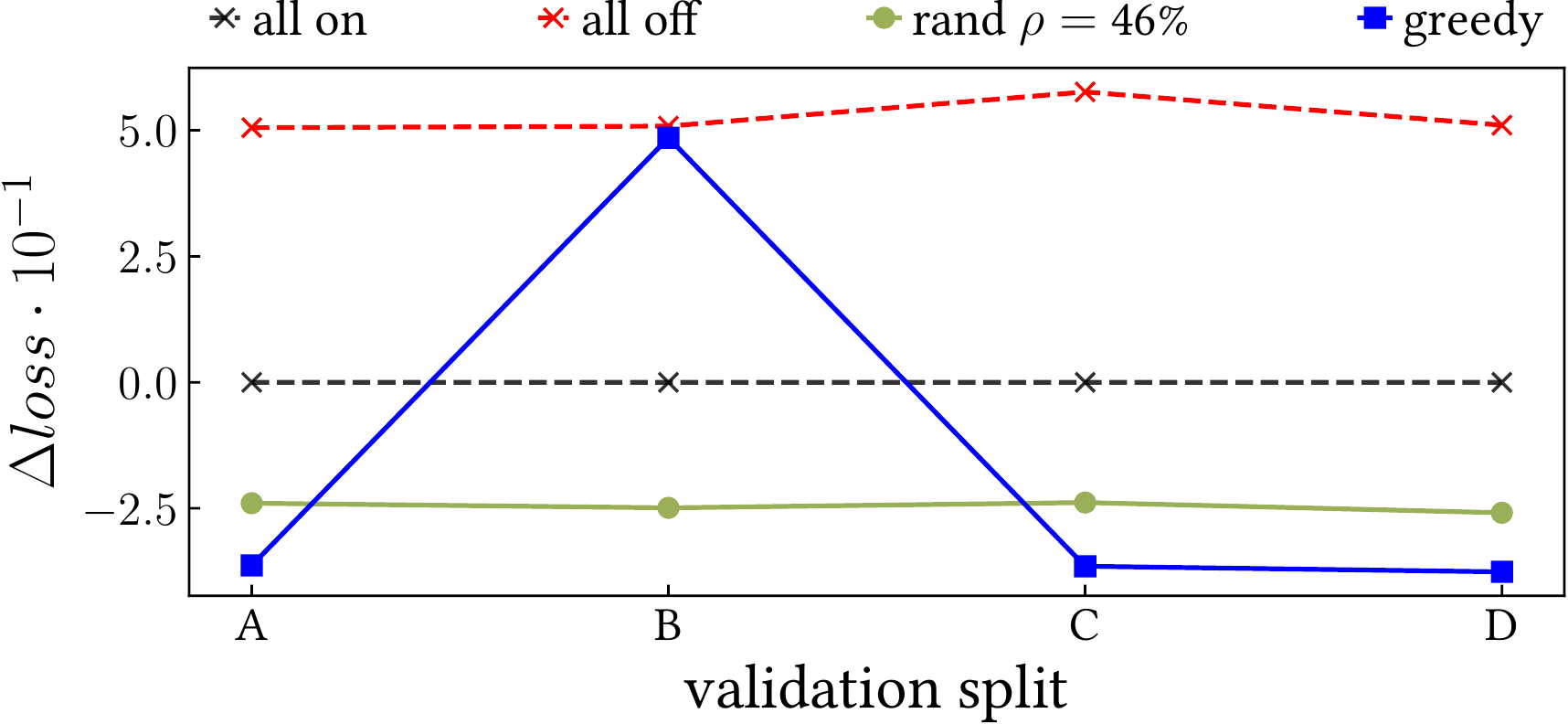}
    		\vspace{-0.6cm}
    		\caption{Greedy expansion against baselines in \textbf{D1}.}
    		\label{fig:dataseta_greedy}
	    \end{figure}
	    
	    When compared against the baselines, the outcomes vary. For \textbf{D1}, the greedy expansion was superior to the baselines except for the second fold, where it failed to met the constraints (Figure~\ref{fig:dataseta_greedy}). In the other three folds, greedy expansion was able to remove $\approx 75\%$ of the rules and reduce alerts. For \textbf{D2}, however, the greedy expansion was worse than the baselines in two of the four folds, as it did not respect the constraints (Supplementary Figure~S\ref{fig:datasetb_greedy}).
	    
\begin{figure}
    		\centering
    		\includegraphics[width=1.0\linewidth]{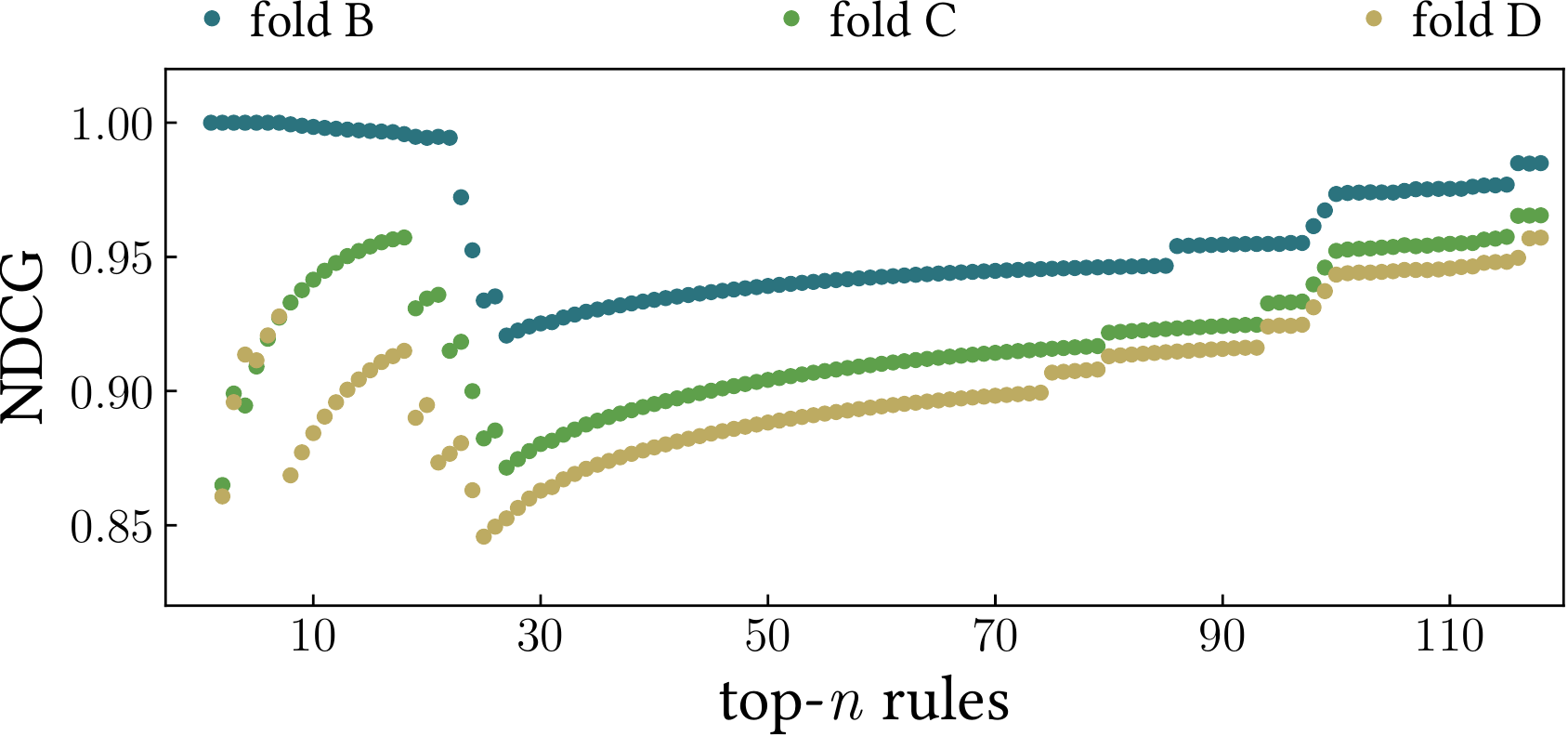}
    		\vspace{-0.6cm}
    		\caption{Greedy expansion rule order consistency in \textbf{D1}.}
    		\label{fig:dataseta_greedy_consistency}
    		\vspace{-0.3cm}
	    \end{figure}

	    
	    We also evaluate the consistency between rules across folds. Recall that the greedy expansion obtains an ordered list of rules sequentially by importance. We compare the ordered lists across folds and compute their normalized discounted cumulative gain (NDCG) in Figure~\ref{fig:dataseta_greedy_consistency}. We show results of the first fold of \textbf{D1} compared with the other folds. We observe that rules are consistent across folds (NDCG values are consistently $>0.7$), but the NDCG line drops (e.g., important rules in fold A are more similar to important rules in fold B than in fold C). We observe similar behavior in \textbf{D2} (omitted for brevity).
	    
	    
	    \subsubsection{Genetic programming results}
	    \label{sec:genetic_real}
	 
	    We evaluate how the genetic programming method (Section~\ref{sec:geneticprogramming}) improves fraud detection. Since our datasets are very big, we can not perform a grid search on all parameters. Thus, we have a three phase process.
	    
	    First, we find a good set of default parameters. For this purpose, we set 
	    $\psi = 100$ 
	    and do grid search on $\alpha$ and $\rho$. We do $n = 10000$ evaluations by default, i.e., for $\psi = 100$, then $r = 100$ runs. 
	    We perform a grid search on $\alpha \in [2\%, 5\%, 10\%, 20\%]$ and $\rho \in [0\%, 2\%, 5\%, 10\%]$. For \textbf{D1}, we find that $\rho = 10\%$ outperforms the baselines across datasplits 
	    and that random search takes longer to achieve similar losses (e.g., for fold A; Supplementary Figure~S\ref{fig:dataseta_genetic_mutation_survivors_evolution}). The overall best parameters were found to be $\rho = 10\%$, $\alpha = 5\%$, and $\rho = 5\%$.
	        
\begin{figure}[t]
    		\centering
    		\includegraphics[width=1.0\linewidth]{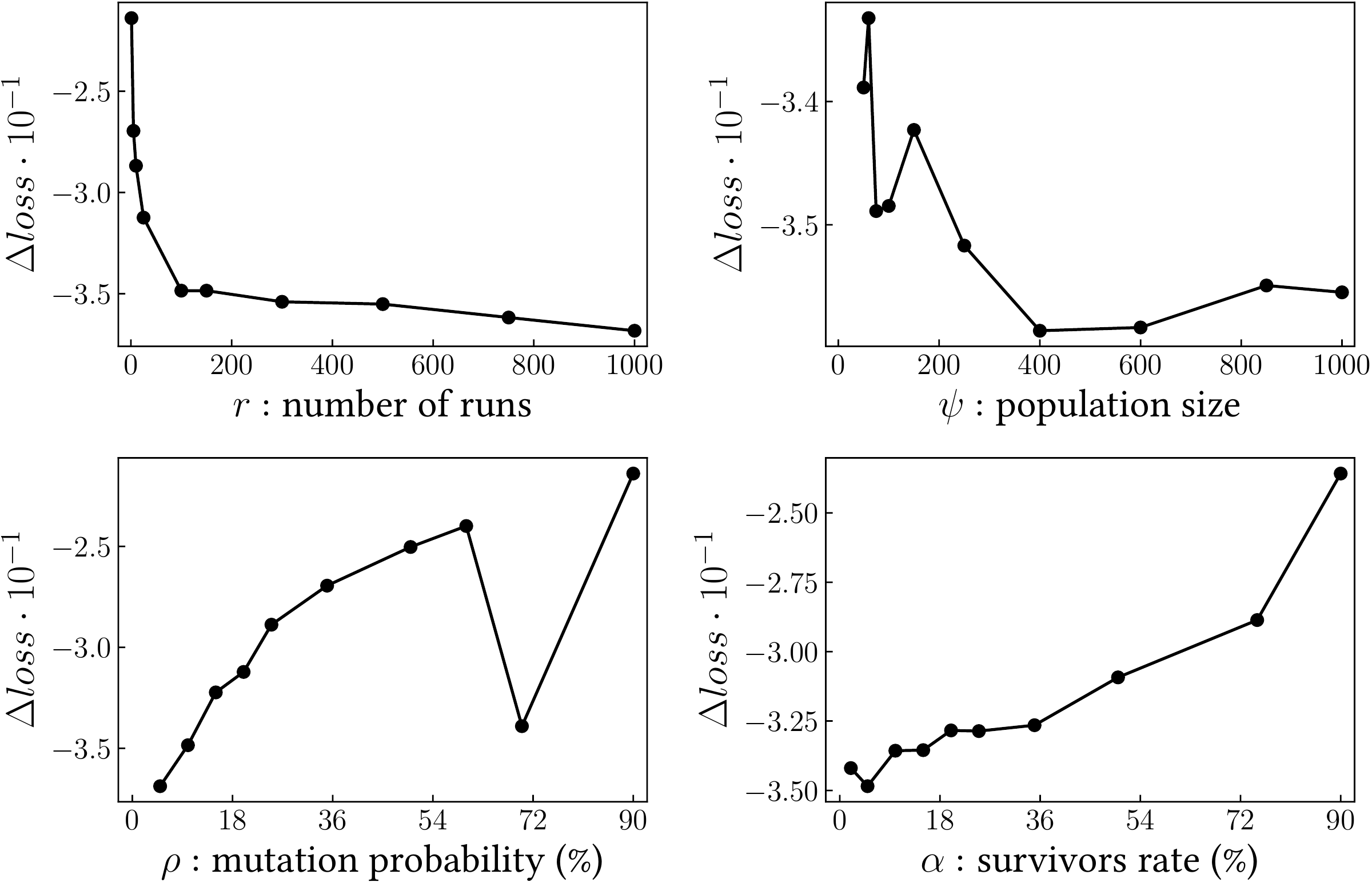}
    		\caption{Genetic: influence of $r$, $\psi$, $\rho$, $\alpha$ on fold A of \textbf{D1}.}
    		\label{fig:dataseta_genetic_foldA_phase2}
    		\vspace{-0.3cm}
	    \end{figure}

	    Secondly, we study how each parameter influences the loss. For this purpose, we vary only one parameter at a time and keep the others at the default values. Since parameters  $r$, $\psi$, $\rho$, $\alpha$ are ordinal, we try 10 different values for each and see how increasing each parameter individually influences the loss. Figure~\ref{fig:dataseta_genetic_foldA_phase2} shows results for fold A of \textbf{D1}. 
	    We observe that, in general, increasing $\rho$ and $\alpha$ makes the performance worse; however the best $\alpha$ is 10\%, thus, keeping some of the best individual configurations is important. We also observe that $r$ influences the loss much more than $\psi$ (e.g., both $(r=100, \psi=10$) and $(r=10, \psi = 100)$ perform 1000 rule evaluations, but the first one leads to lower losses). Typically, the loss improves as you increase $r$ and $\psi$, but it plateaus relatively quickly for both (i.e., $r$ at 300, $\psi$ at 400). Similar conclusions hold for \textbf{D2}
	    (omitted for space concerns).
	    We did not observe gains in changing rule priorities during genetic optimization. 

	     Finally, we measure ARMS performance on the test sets. We compare ARMS using genetic programming against the baselines and ARMS using greedy search. To do this, we evaluate the rules deactivations suggested by ARMS (trained on the train sets and evaluated in the validation sets) on the respective test sets of each fold. For \textbf{D1}, we evaluate the best rule configuration found by ARMS using $r=1000$, $\psi = 250$, $\alpha = 5\%$, $\rho = 5\%$
	     , and no priority shuffling. For \textbf{D2}, we evaluate the best rule configuration found by ARMS using $r=1000$, $\psi = 150$, $\alpha = 20\%$, $\rho = 5\%$
	     , and no priority shuffling. 
	    
	    
	    	    \begin{figure}[b]
    		\centering
    		\includegraphics[width=1.0\linewidth]{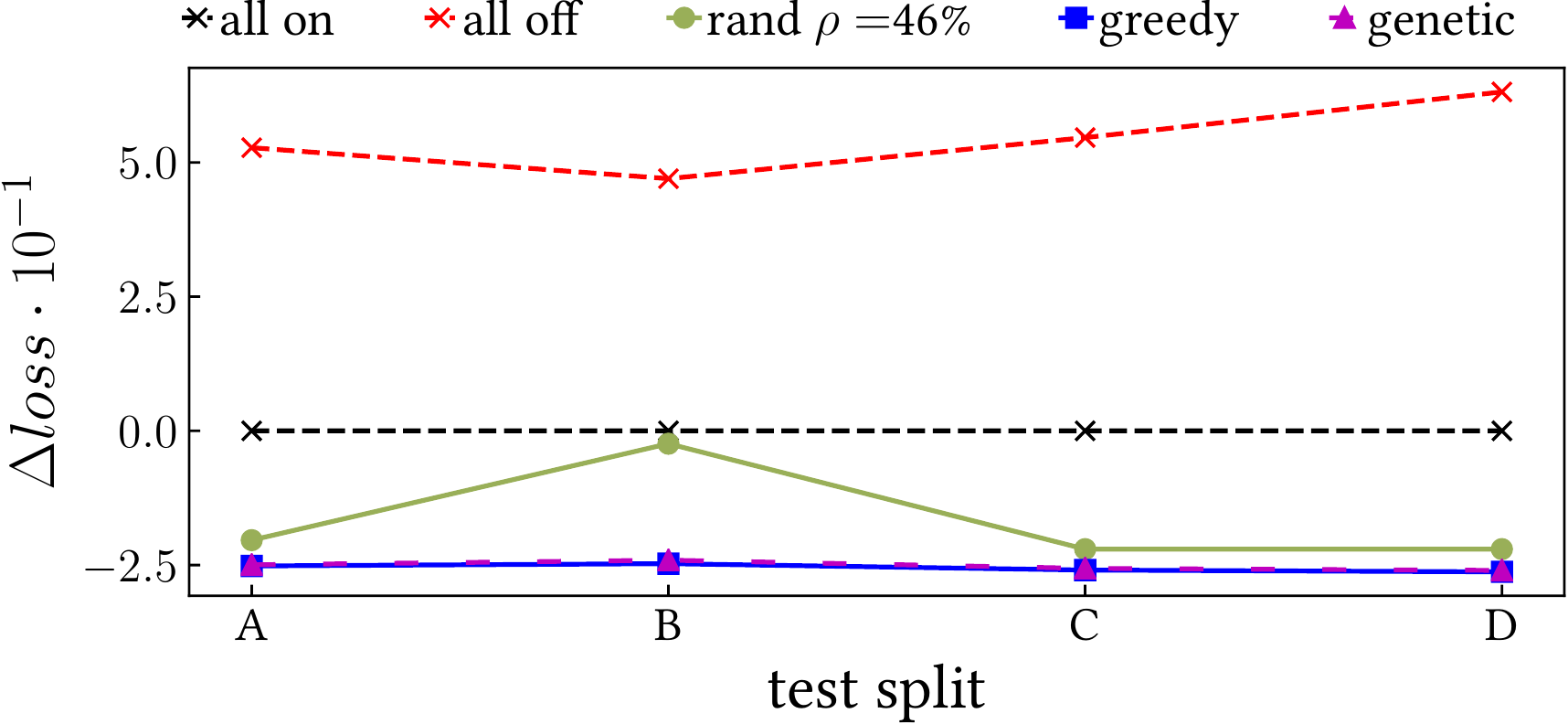}
    		\caption{Performance of ARMS on the test sets of \textbf{D1}.}
    		\label{fig:dataseta_test}
	    \end{figure}
	    
	    \begin{figure}[b]
    		\centering
    		\includegraphics[width=1.0\linewidth]{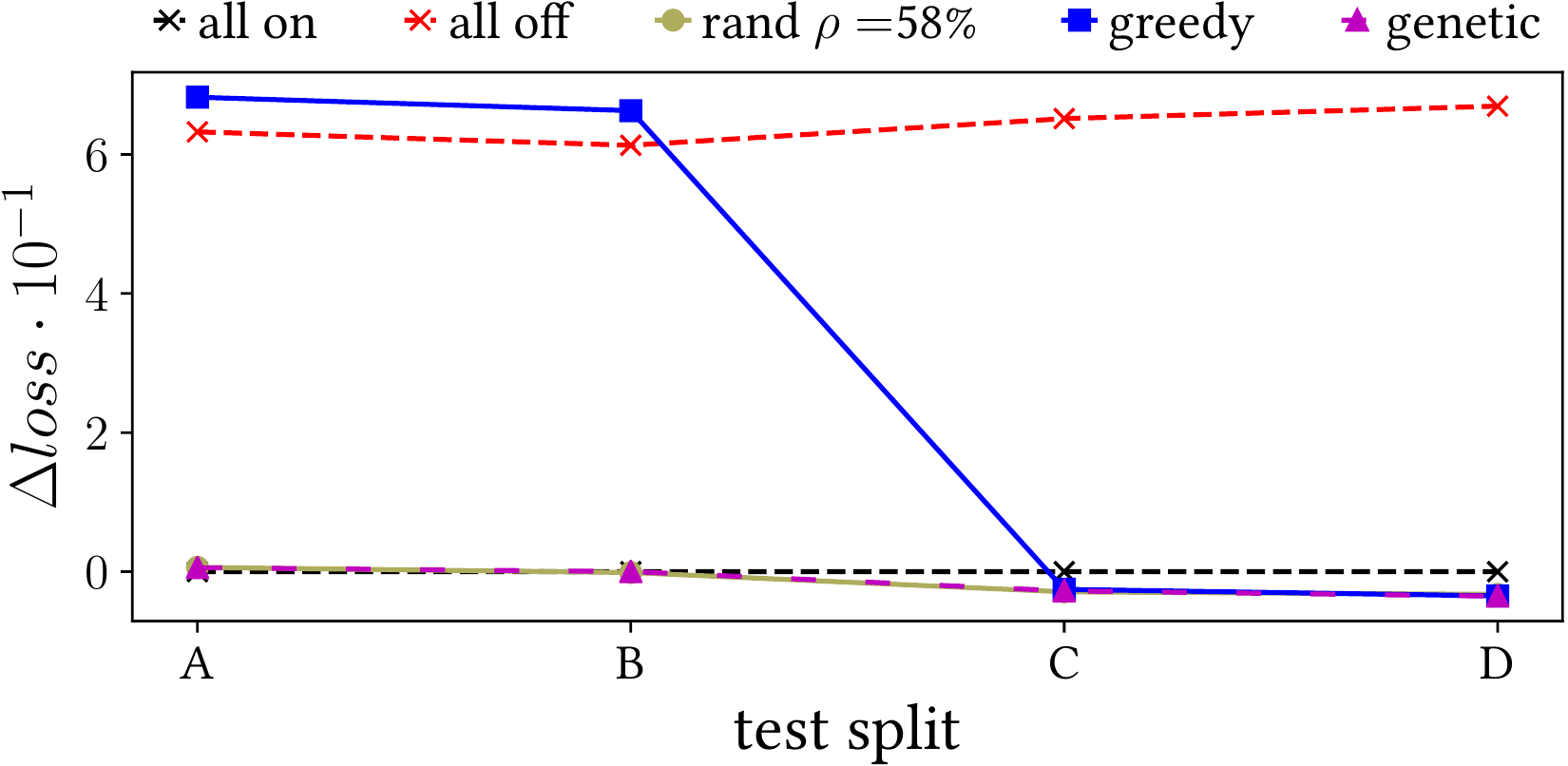}
    		\caption{Performance of ARMS on the test sets of \textbf{D2}.}
    		\label{fig:datasetb_test}
	    \end{figure}
	    
	    For \textbf{D1}, we observe that greedy and genetic optimization performed similarly  and better than random search with $\rho = 46\%$ 
	    (Figure~\ref{fig:dataseta_test}). For \textbf{D2}, we observe that random search and the genetic programming approaches perform similarly; the greedy method fails to comply to the constraints in two of the four folds (Figure~\ref{fig:datasetb_test}).
	    
	    \begin{table*}[t]
	    \setlength{\tabcolsep}{4pt}
	        \small
	        \caption{ARMS consistency results (i.e., across folds). We highlight in bold the lowest loss for each fold. }
	        \label{tab:arms_consistency}
            \begin{minipage}{.24\linewidth}
              \centering
                \begin{tabular}{c|cccc}
                    &  \textbf{A} & \textbf{B} & \textbf{C} & \textbf{D} \\ \hline
                    \textbf{A} & 1 & 0.930 & 0.902 & 0.826 \\
                    \textbf{B} & -- & 1 & 0.950 & 0.820 \\
                    \textbf{C} & -- & -- & 1 & 0.829 \\
                    \textbf{D} & -- & -- & -- & 1 \\
                    \multicolumn{1}{c}{} &  &  &  &  \\
                    \multicolumn{5}{c}{\textbf{(a)} Jaccard of removed rules (\textbf{D1}).}
                \end{tabular}
            \end{minipage}%
            \begin{minipage}{.235\linewidth}
              \centering
                \begin{tabular}{c|cccc}
                     &  \textbf{A} & \textbf{B} & \textbf{C} & \textbf{D} \\ \hline
                    \textbf{A} & 1 & 0.789 & 0.696 & 0.636 \\
                    \textbf{B} & -- & 1 & 0.773 & 0.565  \\
                    \textbf{C} & -- & -- & 1 & 0.708 \\
                    \textbf{D} & -- & -- & -- & 1 \\
                    \multicolumn{1}{c}{} &  &  &  &  \\
                    \multicolumn{5}{c}{\textbf{(b)} Jaccard of removed rules (\textbf{D2}).}
                \end{tabular}
            \end{minipage} 
            \begin{minipage}{.24\linewidth}
              \centering
                \begin{tabular}{c|cccc}
                    &  \textbf{A} & \textbf{B} & \textbf{C} & \textbf{D} \\ \hline
                    \textbf{A} & \textbf{0.275} & \textbf{0.344} & 0.274 & 0.273 \\
                    \textbf{B} & --    & 0.348 & 0.277 & 0.275 \\
                    \textbf{C} & --    & --    & \textbf{0.268} & 0.267 \\
                    \textbf{D} & --    & --    & --    & \textbf{0.264} \\
                    \multicolumn{1}{c}{} &  &  &  &  \\
                    \multicolumn{5}{c}{\textbf{(c)} Loss on future folds (\textbf{D1}).}
                \end{tabular}
            \end{minipage}%
            \begin{minipage}{.264\linewidth}
              \centering
                \begin{tabular}{c|cccc}
                    &  \textbf{A} & \textbf{B} & \textbf{C} & \textbf{D} \\ \hline
                    \textbf{A} & \textbf{-0.626} & \textbf{-0.613} & -0.651 & -0.662 \\
                    \textbf{B} & -- & -0.612 & -0.651 & -0.662 \\
                    \textbf{C} & -- & -- & \textbf{-0.678} & -0.696 \\
                    \textbf{D} & -- & -- & -- & \textbf{-0.704} \\
                    \multicolumn{1}{c}{} &  &  &  &  \\
                    \multicolumn{5}{c}{\textbf{(d)} Loss on future folds (\textbf{D2}).}
                \end{tabular}
            \end{minipage} 
            
        \end{table*}
	    
	    In order to check the consistency of ARMS across data folds, we measure the Jaccard similarity~\cite{urbani1980statistical} of the deactivated rules suggested by ARMS in different splits. We see that the Jaccard is higher for \textbf{D1} than \textbf{D2} (Table~\ref{tab:arms_consistency} (a)-(b)). The fact that \textbf{D2} rule set changes across folds obviously leads to intrinsically lower values (i.e., regardless of what ARMS deactivates). We also evaluate systems trained on a given fold in more recent folds (e.g., we train ARMS on fold A and evaluate it the test set of A, B, C and D). We observe that systems trained on older folds have good performance on more recent test sets (Table~\ref{tab:arms_consistency} (c)-(d)).



	\subsubsection{Summary}
	    
    We evaluated ARMS on two big online merchants. For \textbf{D1}, ARMS using genetic programming (or greedy expansion) was able to remove $\approx$ 50\% of the original 193 rules, while maintaining the original system performance (i.e., keeping 95\% of the original recall). Thus, ARMS was able to improve the original system (\textbf{h1}). We also saw that results are stable across data-splits (\textbf{h3}). We did not see gains of using priority shuffling (\textbf{h2}). For D2, we observed that ARMS was able to remove $\approx$ 80\% of the system rules while maintaining the original system performance (i.e., keeping a low FPR). Thus, ARMS improved the original system (\textbf{h1}). Similar to \textbf{D1}, we found  evidence supporting (\textbf{h3}) but not (\textbf{h2}).
    
	 \subsubsection{Discussion}
	 Real-world transaction data sets for fraud detection pose several challenges. Auto-declines lead to unreliable labels, and thus we cannot verify if a system positive is a true positive, meaning that decline rules cannot be evaluated unless an analyst verifies auto-declines. In practice this is difficult because fraud analysts' time is a very limited resource. The two systems that we chose are also particularly hard to optimize since they have been in production for years and have been manually tuned by data scientists. Finally, we evaluated ARMS' performance on past transactions and did not measure its performance in production. We think that putting ARMS in production and continuously optimizing the rules system could lead to better results.

	\section{Conclusion}
	\label{sec:conclusions}
	
	We have proposed ARMS, a framework that optimizes rules systems using search heuristics, namely random search, greedy expansion, and genetic programming. To the best of our knowledge, ARMS is the first to (1) handle different rule priorities and actions, (2) address blacklists side effects, and (3) optimize user-defined functions. These components are essential in real-world fraud detection systems. Our results in real-world clients demonstrate that ARMS is capable of maintaining the original system's performance while greatly reducing the number of rules (between 50\% and 80\%, in our experiments) and minimizing other metrics (e.g., alert rate).
	
	Currently we are adding a rules suggestions module to ARMS, which is beyond the scope of this paper. In the future we also plan to incorporate a module to simultaneously tune the rules and the machine learning model threshold.
	

	\begin{acks}
	  We want to thank the other members of Feedzai's research team, who always gave insightful suggestions. In particular, we want to give special thanks to Marco Sampaio, for reviewing the paper internally,  and Patr\'icia Rodrigues, for starting ARMS.
	\end{acks}
	
	\subsection*{Note on reproducibility}
	
	We make available a binary of ARMS, the synthetic data  described in Section 4.1 (as well as the script used to generate it), and all the necessary steps to reproduce our results from Section~4.3 at \url{https://github.com/feedzai/research-arms}. For privacy compliance, we can not share our clients data sets.

	\bibliographystyle{ACM-Reference-Format}
	\bibliography{arms_bib.bib} 
	
	\pagebreak
	
	\setcounter{section}{0}
	\setcounter{algorithm}{0}
	\setcounter{figure}{0}
	\setcounter{table}{0}
	\renewcommand{\thesection}{\Alph{section}}
	\section{Supplementary materials}

\makeatletter
\newenvironment{supalgorithm}[1][htb]{%
	\renewcommand{\ALG@name}{Algorithm}
	\begin{algorithm}[#1]%
	}{\end{algorithm}}
\makeatother

\makeatletter
\newenvironment{supfigure}[1][htb]{%
	\renewcommand{\figurename}{Figure}
	\begin{figure}[#1]%
	}{\end{figure}}
\makeatother

\makeatletter
\newenvironment{suptable}[1][htb]{%
	\renewcommand{\tablename}{Table}
	\begin{table}[#1]%
	}{\end{table}}
\makeatother

\makeatletter
\renewcommand{\fnum@figure}{\figurename~S\thefigure}
\makeatother

\makeatletter
\renewcommand{\fnum@table}{\tablename~S\thetable}
\makeatother

\makeatletter
\renewcommand{\fnum@algorithm}{\ALG@name~S\thealgorithm}
\makeatother


\subsection{Synthetic data}
\label{sup_sec:synthetic}

The set of rules comprises 8 accept rules, 30 review rules, and 60 decline rules. The support of the accept rules was sampled from a Gaussian distribution $\mathcal{N}(45000, 22500^2)$, while the support of the review and decline rules was sampled from $\mathcal{N}(22.5, 225.0^2)$. The NPV of accept rules was sampled from $\mathcal{N}(0.75,0.20^2)$, while the precision of the alert and decline rules was sampled from $\mathcal{N}(0.17,0.05^2)$.

Rules have ten possible priorities. Accept rules have priority $p_a \in \{0, 1, 5, 6, 10\}$, alert rules have priority $p_l \in \{2, 4, 7, 9\}$, and decline rules have priority $p_d \in \{3, 8\}$.

\subsection{Synthetic data parameter tuning}\label{sec:hyperparams}

\subsubsection{Random search}

We use 16 mutation probabilities, i.e., $\rho \in [4\%, 94\%]$, in intervals of 4\%, i.e., $\rho = 4\%$, $\rho = 8\%$, ... 

\subsubsection{Genetic programming}

We use three mutation probabilities, i.e., $\rho = 10\%$, $\rho = 20\%$, and $\rho = 30\%$. We use two population sizes, i.e., $\psi = 20$ and $\psi = 30$. Finally, we use two survivors fractions, i.e., $\alpha = 2\%$ and $\alpha = 5\%$.

\subsection{Real-world datasets}
\label{sup_section:realworlddatasets}

\subsubsection{D1}

The client has 198 rules, with one of three possible actions: accept, alert, and decline. Out of the 198 rules, 30 of them are accept rules, 89 are alert rules, and 79 are decline rules. Accept rules have four different priority levels $p_a \in \{1, 8, 10, 15\}$, alert rules have two $p_a \in \{5, 11\}$, and decline rules have three $p_d \in \{6, 9, 12\}$. If no rules are triggered, the default action is to accept the transaction. 

The dataset contains few validated fraud, i.e., of the declined (by the model/rules) and fraudulent population of transactions, only a small portion was validated by analysts or via chargeback.

We note that decline rules and auto-declined transactions are ignored in the train and validation datasets. We make this choice because decline rules can not be validated. However, when we measure performance in the test set, decline rules are included in order to make results directly comparable to the results obtained in production.


We do temporal cross validation (TCV) with four folds and each set has one month of data.

\subsubsection{D2}

Unlike \textbf{D1}, which has the same activated rules for the whole period, in D2 the rules changed. During the seven months period a total of 13 rules were added, while some were removed, increasing the number of rules in the set from the original 77 to 90. 

Rules have one of three outcomes: accept, alert, and alert\&decline (this means that most auto-declined are verified, unlike in \textbf{D1}). From those, 6 are accept rules, 48 are alert rules, and 36 are alert\&decline rules. Accept rules have priority $p_a \in \{0, 5, 10\}$, alert rules have priority $p_l = 1$, and alert\&decline rules have priority $p_d \in \{2, 4, 8\}$. Three of the decline rules are blacklist checker rules, and all 36 alert\&decline rules are blacklist updater rules.

Since \textbf{D2} has a high ratio of validated fraud, all rules are optimized by ARMS, however the auto-decline transactions are not used during the training process, but are present in the test set in order to make results directly comparable to the results obtained in production.

We do temporal cross validation (TCV) with four folds and each set has one month of data.

\subsection{Supplementary Algorithms and Figures}

\begin{algorithm}[h]
    \footnotesize
		\caption{Blacklist propagation.} 
		\label{alg:blacklists}
		\begin{algorithmic}[1]
			\Function{computeBlacklistDependencies}{$\mathbf{R}$, $\mathbf{X}$, $\mathcal{X}$, $\mathcal{B}$}
			\State $\mathbf{BL} \gets \{\}$ 
			\State $\mathbf{BD} \gets \{\}$ 
			
			\ForAll{$\mathbf{x} \in \mathbf{X}$}
			\ForAll{$R_j \in \mathcal{B}^u$}
			\If{$r_j \neq -1$} 
			\ForAll{$X_l \in \mathcal{X}$ that $R_j$ blacklists}
			
			\State $\mathbf{BL}[(R_j, X_l:x_l)].\Call{append}{[\mathbf{x}.time, +\infty]}$
			
			\ForAll{$R_q \in \mathcal{B}^c$ that checks $X_l$}
			\State $\mathbf{BD}[\mathbf{x}]$.\Call{add}{$R_j \prec R_q$}
			\EndFor
			\EndFor
			\EndIf
			\If{$r_j = -1$} 
			\ForAll{$X_l \in \mathcal{X}$ that $R_j$ can blacklist}
			
			\If{$\mathbf{x}.time$ is in any $\mathbf{BL}[(R_j, X_l:x_l)]$}
			\State $r_j \gets p_j$
			\EndIf
			\EndFor
			\EndIf
			\EndFor
			\ForAll{$\{x_l \in \mathbf{x} \mid x_l$  is in any active blacklist $\}$}
			%
			%
			\If{$(\not \exists R_q \in \mathcal{B}^c \mid p_q \neq -1)$ }
			
			\ForAll{$\{R_j \in \mathcal{B}^u\ \mid (R_j, X_l : x_l) \in \mathbf{BL}\}$}
			\State $\mathbf{BL}[(R_j, X_l:x_l)]$.\Call{last}{{}} $\gets [\_, \mathbf{x}.time]$
			\EndFor
			\EndIf
			\EndFor
			
			\ForAll{$R_q \in \mathcal{B}^c$} 
			\If{$r_q \neq -1$ \textbf{ and} 
					$|\{(R_i \prec R_q) \in \mathbf{BD}[\mathbf{x}]\ \mid R_i \in \mathcal{B}^u\}| = 0$}
			\State $\mathbf{BD}[\mathbf{x}]$.\Call{add}{$R_q \prec R_q$)}
			\EndIf
			\EndFor
			\EndFor
			
			\State \Return $\mathbf{BD}$				
			\EndFunction
			
		\end{algorithmic}
	\end{algorithm}

	\begin{algorithm}[!h]
    \footnotesize
		\caption{Random search optimization.} 
		\label{alg:random}
		\begin{flushleft}
			$\theta$: \{ rule shutoff probability $\rho$, rule priority shuffle probability $\gamma$ \}
		\end{flushleft}
		\begin{algorithmic}[1]
			\Function{Random.optimize}{$\mathbf{X}$, $\mathbf{R}$, $\boldsymbol\ell$, $\mathbf{p}$, $a$, $\mathbf{BD}$, $\lambda$, $\Omega^{1}$, $\theta$}
			\State $\mathbf{p^{best}} \gets \mathbf{p}$
			\State $\Omega^{best} \gets \Omega^1$
			\While{\Call{stoppingCriteriaNotMet}{{}}}
			\State $\mathbf{p^{rand}} \gets \mathbf{p}$
			\ForAll{$p_i \in \mathbf{p^{rand}}$}
			\State \textbf{with} $\gamma \%$ probability, \textbf{do}:
			\State \quad $p_i \gets$ \Call{randomPriorityShuffle}{$p_i$, $a$}
			\State \textbf{with} $\rho \%$ probability, \textbf{do}:
			\State \quad $p_i \gets -1$
			\EndFor 
			\State $\Omega^{rand} \gets$ \Call{evaluate}{$\mathbf{X}$, $\mathbf{R}$, $\boldsymbol\ell$, $\mathbf{p^{rand}}$, $a$, $\mathcal{B}$, $\mathbf{BD}$, $\lambda$}
			\If{$\Omega^{rand}_{loss} < \Omega^{best}_{loss}$}
			\State $\Omega^{best} \gets \Omega^{rand}$
			\State $\mathbf{p^{best}} \gets \mathbf{p^{rand}}$
			\EndIf
			\EndWhile
			\State \Return ($\mathbf{p^{best}}, \Omega^{best}$)
			\EndFunction
			
		\end{algorithmic}
	\end{algorithm}

	\begin{algorithm}[!h]
    \footnotesize
		\caption{Greedy expansion optimization.} 
		\label{alg:greedy}
		\begin{flushleft}
			$\theta$: \{ backtracking $bt \in \{true, false\}$ \}
		\end{flushleft}
		\begin{algorithmic}[1]
			\Function{Greedy.optimize}{$\mathbf{X}$, $\mathbf{R}$, $\boldsymbol\ell$, $\mathbf{p}$, a, $\mathbf{BD}$, $\lambda$, $\Omega^{1}$, $\theta$}
			\State $\mathbf{p^{best}} \gets \mathbf{p}$
			\State $\Omega^{best} \gets \Omega^1$
			\State $\mathbf{p^{keep}} \gets (-1, ..., -1)$
			\State $\mathbf{p^{greedy}} \gets (-1, ..., -1)$
			\State $Q \gets \emptyset$
			\While{$|Q| < |\mathcal{R}| ${ \textbf{and}} \Call{stoppingCriteriaNotMet}{{}}}
			\State $R_{keep} \gets ${ None}
			\State $\Omega^{keep} \gets +\infty$
			\ForAll{$\{R_j \in \mathcal{R} \mid R_j \not \in Q\}$}
			\State $p^{greedy}_j \gets p_j$
			\State $\Omega^{greedy} \gets$ \Call{evaluate}{$\mathbf{X}$, $\mathbf{R}$, $\boldsymbol\ell$, $\mathbf{p^{greedy}}$, $a$, $\mathcal{B}$, $\mathbf{BD}$, $\lambda$}
			\If{$\Omega^{greedy}_{loss} < \Omega^{keep}_{loss}$}
			\State $R_{keep} \gets R_j$ 
			\State $\Omega^{keep} \gets \Omega^{greedy}$
			\State $\mathbf{p^{keep}} \gets \mathbf{p^{greedy}}$
			\EndIf
			\State $p^{greedy}_j \gets -1$
			\EndFor
			\State $Q$.\Call{add}{$R_{keep}$}
			\If{$\Omega^{keep}_{loss} < \Omega^{best}_{loss}$}
			\State $\Omega^{best} \gets \Omega^{keep}$
			\State $\mathbf{p^{best}} \gets \mathbf{p^{keep}}$
			\EndIf
			\If{$bt$ is $true$\textbf{ and } \Call{isBacktrackingTime}{{}}}
			\State run greedy contraction to remove $l$ rules, $l < |Q|$
			\EndIf
			\EndWhile
			\State \Return ($\mathbf{p^{best}}, \Omega^{best}$)
			\EndFunction
			
		\end{algorithmic}
	\end{algorithm}	

	\begin{algorithm}[!h]
    \footnotesize
		\caption{Genetic programming optimization.} 
		\label{alg:genetic}
		\begin{flushleft}
			$\theta$: \{ Population size $\psi$, survivors fraction $\alpha$, mutation probability $\rho$
			\}
		\end{flushleft}
		\begin{algorithmic}[1]
			\Function{Genetic.optimize}{$\mathbf{X}$, $\mathbf{R}$, $\boldsymbol\ell$, $\mathbf{p}$, $a$, $\mathbf{BD}$, $\lambda$, $\Omega^{1}$, $\theta$}
			\State $\mathbf{p^{best}} \gets \mathbf{p}$
			\State $\Omega^{best} \gets \Omega^1$
			\State $\mathbf{P} \gets $ 
			\Call{generateInitialPopulation}{$\mathbf{R}$, $\mathbf{p}$, $\psi$, $\rho$}
			\While{\Call{stoppingCriteriaNotMet}{{}}}
			\State $(\mathbf{P^{=}}, \mathbf{P^{-}}) \gets $ \Call{evaluatePopulation}{$\mathbf{P}$, $\alpha$}
			\State $\mathbf{P^{+}} \gets $ \Call{mutateAndCrossover}{$\mathbf{P^{=}}$, $\alpha$, $\psi$, $\rho$}
			\State $\mathbf{P} \gets \{\mathbf{P^=}, \mathbf{P^+}\}$
			\EndWhile
			\State $(\mathbf{P^{=}}, \mathbf{P^{-}}) \gets $ \Call{evaluatePopulation}{$\mathbf{P}$}
			\State $\mathbf{p^{best}} \gets \mathbf{P^{=}_1}$
			\State $\Omega^{best} \gets$ \Call{evaluate}{$\mathbf{X}$, $\mathbf{R}$, $\boldsymbol\ell$, $\mathbf{p^{best}}$, $a$, $\mathcal{B}$, $\mathbf{BD}$, $\lambda$}
			\State \Return $(\mathbf{p^{best}}, \Omega^{best})$
			\EndFunction
			
			\vspace{0.2cm}
			\Function{generateInitialPopulation}{$\mathbf{R}$, $\mathbf{p}$, $\psi$, $\rho$}
			\State $\mathbf{P} \gets \emptyset$
			\For{$i \in [0, \psi[$}
			\State $\mathbf{p'} \gets \mathbf{p}$
			\ForAll{$p'_j \in \mathbf{p'}$}
			\State \textbf{with} $\rho \%$ probability, \textbf{do}:
			\State \quad $p'_j \gets -1$
			\EndFor
			\State $\mathbf{P[i]} \gets \mathbf{p'}$
			\EndFor
			\State \Return $\mathbf{P}$	
			\EndFunction
			
			
			\vspace{0.2cm}
			\Function{mutateAndCrossover}{$\mathbf{P^{=}}$, $\alpha$, $\psi$, $\rho$}
			\State $\mathbf{P^+} \gets \emptyset$
			\For{$i \in [0, (1-\alpha) * \psi[$}
			\State $\mathbf{p^{mother}} \gets $ \Call{getRandomVector}{$\mathbf{P^=}$}
			\State $\mathbf{p^{father}} \gets $ \Call{getRandomVector}{$\mathbf{P^=}$}
			\State $\mathbf{p^{child}} \gets \mathbf{p^{mother}}$
			\ForAll{$p_j^{child} \in \mathbf{p^{child}}$}
			\State \textbf{with} $50 \%$ probability, \textbf{do}:
			\State \quad $p_j^{child} \gets p_j^{father}$
			\EndFor
			\ForAll{$p_j^{child} \in \mathbf{p^{child}}$}
			\State \textbf{with} $\rho \%$ probability, \textbf{do}:
			\State  \quad $p_j^{child} \gets $ \Call{randomPriorityShuffle}{$p_i$, $a$}
			\EndFor
			\State $\mathbf{P^+}$.\Call{add}{$\mathbf{p^{child}}$}
			\EndFor
			\State \Return $\mathbf{P^+}$
			\EndFunction
			
		\end{algorithmic}
	\end{algorithm}

	        \begin{figure}[h]
    		\centering
    		\includegraphics[width=1\linewidth]{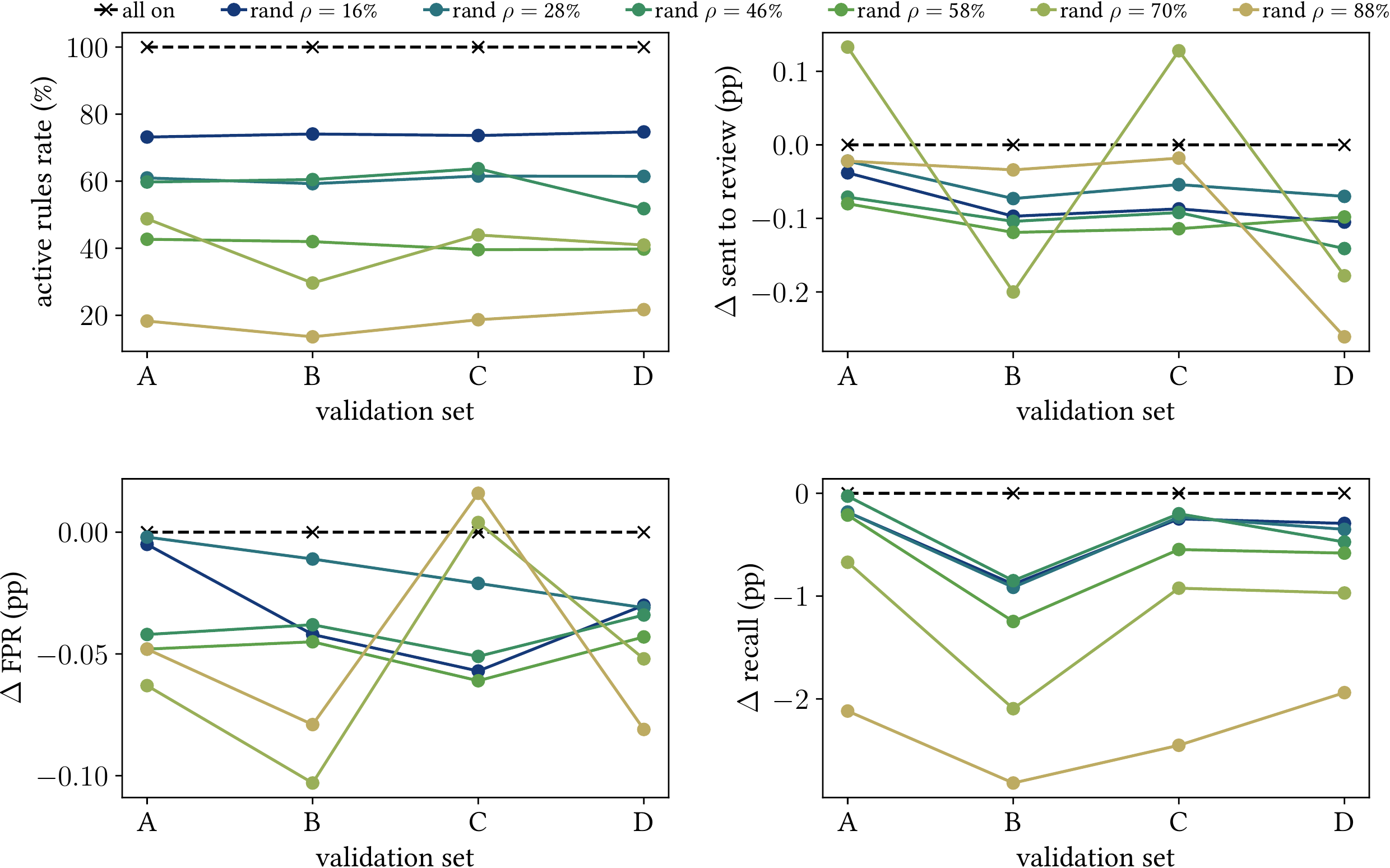}
    		\caption{Baseline metrics comparison in D2.}
    		\label{fig:datasetb_baselines}
	    \end{figure}
	
	\begin{figure}[h]
    		\centering
    		\includegraphics[width=1\linewidth]{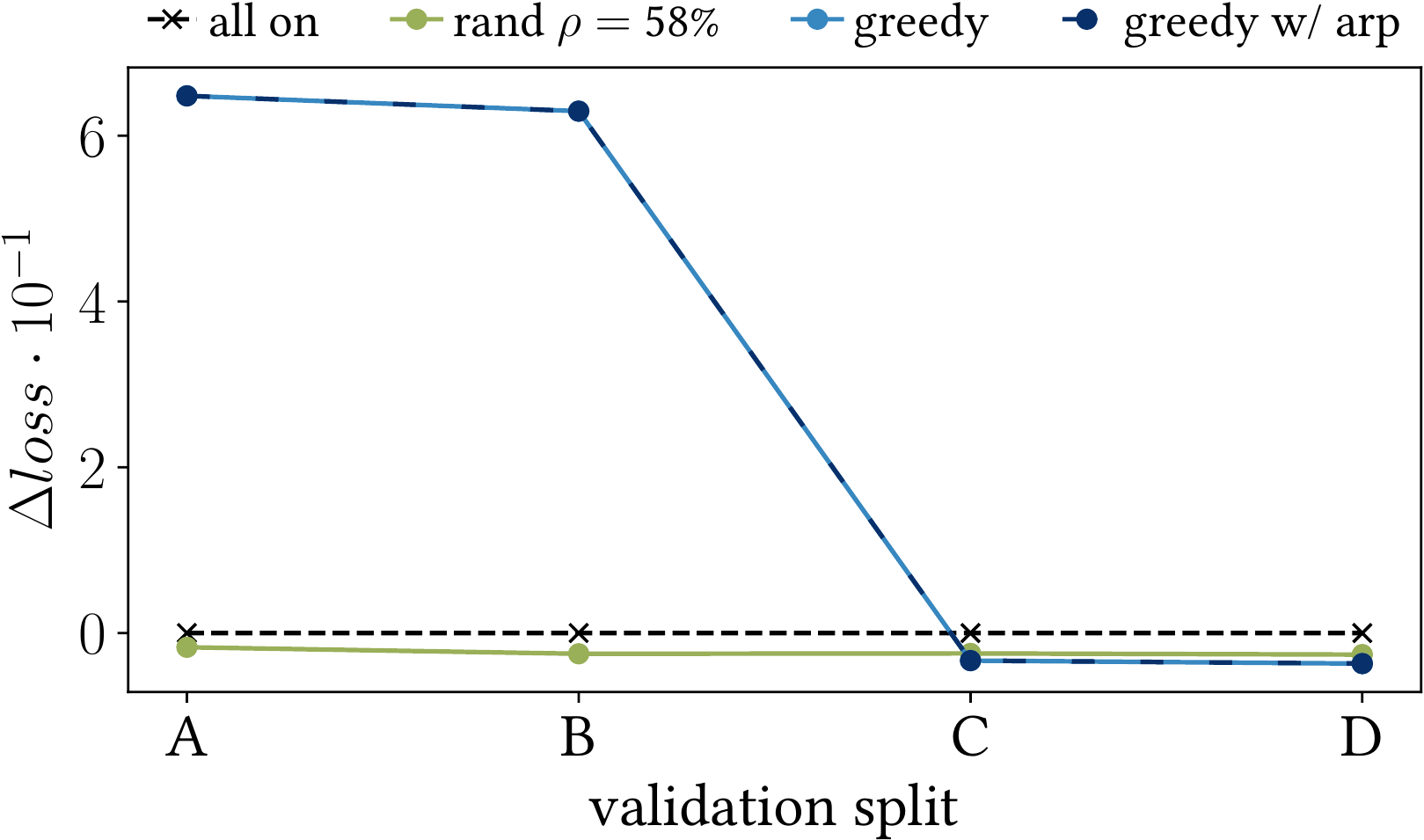}
    		\caption{Greedy expansion results in D2.}
    		\label{fig:datasetb_greedy}
	    \end{figure}
	    
	    	    \begin{figure}[h]
    		\centering
    		\includegraphics[width=1\linewidth]{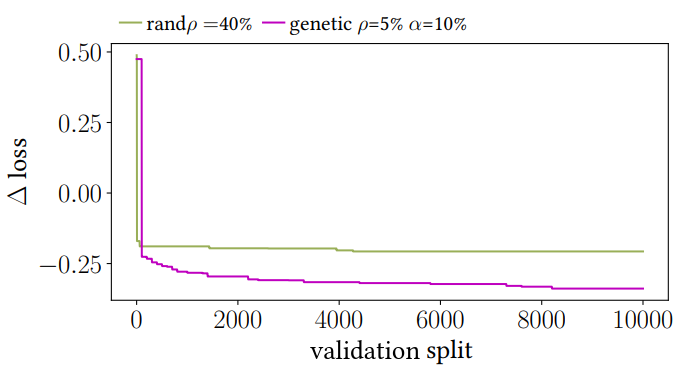}
    		\caption{Genetic programming loss versus random search by number of evaluations in fold A of D1 (zoomed in the first 10000 rule evaluations; the methods  nearly converge eventually)}
    		\label{fig:dataseta_genetic_mutation_survivors_evolution}
	    \end{figure}
	\textbf{}
\end{document}